\documentclass[journal]{IEEEtran}
\usepackage{color}
\usepackage{algorithm}
\usepackage{algorithmic}
\usepackage{amsmath} 
\usepackage{amsthm}
\usepackage{amssymb}
\usepackage{graphicx}
\usepackage{booktabs}
\usepackage{threeparttable}
\usepackage{multirow}
\usepackage{tabularx}
\usepackage{url}
\usepackage{cite}

\newtheorem{mydef}{Definition}
\newtheorem{mytheo}{Theorem}
\newtheorem{remark}{Remark}
\usepackage[normalem]{ulem}
\useunder{\uline}{\ul}{}

\begin{document}

\title{Patch Tracking-based {Streaming} Tensor Ring Completion for Visual Data {Recovery}\thanks{This work was supported by NSF CAREER Award CCF-1552497 and NSF Award CCF-2106339.

Yicong He and George K. Atia are with the Department of Electrical and Computer
Engineering, University of Central Florida, Orlando, FL, 32816, USA. e-mails: Yicong.He@ucf.edu, George.Atia@ucf.edu.}}

\author{Yicong He and George K. Atia, \IEEEmembership{Senior~Member,~IEEE}}

\maketitle

\begin{abstract}
Tensor completion {aims to recover} the missing entries of a partially observed tensor by exploiting its low-rank structure, {and has been applied to visual data recovery}. In applications where the data arrives sequentially such as streaming video completion, the missing entries of the tensor need to be dynamically recovered in a streaming fashion. Traditional streaming tensor completion algorithms treat the entire visual data as a tensor, which may not work satisfactorily when there is a big change in the tensor subspace along the temporal dimension, such as due to strong motion across the video frames. In this paper, we develop a novel patch tracking-based {streaming} tensor ring completion framework for visual data recovery. {Given a newly incoming frame, small patches are tracked from the previous frame. Meanwhile, for each tracked patch, a patch tensor is constructed by stacking similar patches from the new frame. Patch tensors are then completed using a streaming tensor ring completion algorithm, and the incoming frame is recovered using the completed patch tensors.} We propose a new patch tracking strategy that can accurately and efficiently track the patches with missing data. Further, a new {streaming} tensor ring completion algorithm is proposed which can efficiently and accurately update the latent core tensors and complete the missing entries of the patch tensors. Extensive experimental results demonstrate the superior performance of the proposed algorithms compared with both {batch} and streaming state-of-the-art tensor completion methods.
\end{abstract}


\IEEEpeerreviewmaketitle

\section{Introduction}
Multi-way data analysis uses techniques that represent data as multi-dimensional arrays known as tensors. It has drawn increased attention in recent years given its ability to reveal patterns intrinsic to high-order data undetected by other methods by capturing correlations across its different dimensions.  
Such techniques have found numerous applications in machine learning \cite{cohen2016expressive,sidiropoulos2017tensor,makantasis2018tensor,kossaifi2020tensor,jia2021multi}, signal processing \cite{rupp2015tensor,bousse2017tensor,de2021channel} and computer vision \cite{sobral2015online,bengua2017efficient,jiang2017novel,li2018harmonious,liu2019low2}.

Tensor completion, an extension to the matrix completion problem, aims to fill in the missing entries of a partially observed tensor by leveraging its low-rank structure, which stems from the redundancy of the information it represents \cite{gandy2011tensor,liu2012tensor}. For example, a natural multi-channel image or a video sequence can be represented as tensors that exhibit high correlations across the spatial and temporal dimensions. Based on different tensor rank models, several completion algorithms have been proposed, such as CP-based algorithms \cite{kiers2000towards,jain2014provable}, tucker-based algorithms \cite{tucker1966some,kasai2016low,xu2015parallel}, tubal (t-SVD)-based algorithms \cite{zhang2016exact,zhou2017tensor,liu2019low}, tensor train-based algorithms \cite{bengua2017efficient,liu2021simulated} and tensor ring-based algorithms \cite{wang2017efficient,yuan2019tensor,huang2020provable}. Extensive experimental results of these tensor completion algorithms have shown their superior completion performance compared to matrix completion. 

\begin{figure}[tb]
\centering
\includegraphics[width=0.97\linewidth]{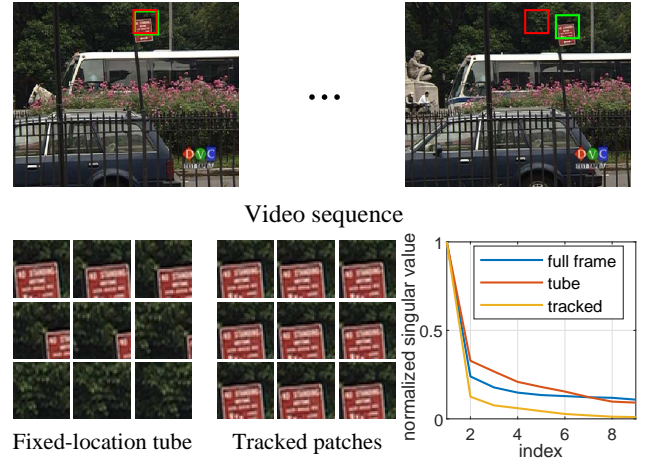}
\caption{Illustration of the patch tracking-based method for the video sequence in the top row. Starting from the first frame, the green patches are tracked along the temporal dimension while the positions of red patches are kept unchanged. The bottom row shows the extracted patches, and the normalized eigenvalues (i.e., each eigenvalue divided by the largest eigenvalue) of the mode-4 unfolding matrix of the tensor formed by stacking the full frames, fixed-location patches (tube), and tracked patches along the fourth (temporal) dimension.}
\label{fig:insight}
\end{figure}

Traditional {batch} tensor completion methods are of limited use in streaming applications wherein the data arrives sequentially, since they are designed to work on the entire tensor data. To mitigate this problem, several methods have been proposed to complete the missing entries of data in a {streaming or online} manner. In \cite{mardani2015subspace}, the authors developed an online tensor completion approach using the CP model and stochastic gradient descent (SGD). The authors in \cite{kasai2016online} proposed an online low-rank subspace tracking by tensor CP decomposition (OLSTEC), which improves the efficiency and accuracy of \cite{mardani2015subspace} by utilizing recursive least squares (RLS) instead of SGD. {Then, \cite{song2017multi} developed a CP-based multi-aspect streaming tensor completion considering the case that the tensor grows along multiple modes.} Further, a tensor rank-one update on the complex Grassmannian (TOUCAN) algorithm is proposed under the t-SVD model \cite{gilman2020online}. Also, a sequential tensor completion (STC) approach was developed for the recovery of internet traffic data under the Tucker tensor model \cite{xie2018accurate}. 

Despite the success of these {streaming} methods in completing the missing entries of sequential data, there are important limitations that remain to be addressed. In particular, all existing methods require that the entire tensor subspace should be stable, that is, the latent tensor factors should be invariant or changing slowly along the temporal dimension. In the presence of large motion between adjacent tensors, the subspace may change considerably, resulting in degraded completion performance. As shown in the first row of Fig.\ref{fig:insight}, some objects (car, road sign) move significantly across different frames. Such large motions increase the rank of the mode-4 unfolding matrix (see bottom right of Fig. \ref{fig:insight}), indicating a big change in the tensor subspace. Thus, existing {streaming} tensor completion algorithms may not accurately complete the objects with large motion.    

In this paper, we introduce a novel patch tracking-based {streaming} tensor ring completion framework for visual data recovery (see Fig.~\ref{fig:PTSTRC}). Rather than performing completion on {the entire incoming frame}, we complete small patch tensors constructed from similar patches extracted and tracked from the frame. As a new frame comes in, similar patches across the temporal domain are tracked using patch matching. For example, as shown in Fig. \ref{fig:insight}, the patches of `road sign' are tracked. Compared with fixed tube and full-frame tensors, the normalized eigenvalues of the mode-4 unfolding matrix of the tracked patches are relatively small, indicating a more stable tensor subspace along the temporal domain. Therefore, the completion performance can be improved substantially when applied to similar tracked patches.

Patch-based methods have been widely applied to visual data processing, such as image and video denoising \cite{dabov2007image,ji2010robust,maggioni2012nonlocal,wen2017joint}. However, patch matching becomes more challenging when the data has missing entries, due to insufficient matching information. 
The authors of \cite{zhang2019nonlocal} address the problem of patch matching with missing entries by interpolating images whose missing pixels are filled using triangle interpolation. An alternative approach is used in \cite{ng2020patched} where the missing pixels are estimated by applying tensor completion before matching. These interpolation or completion methods, however, incur more computational cost and may introduce additional errors to patch matching. The authors in \cite{ding2021tensor} proposed a tube matching approach by adding the information of the temporal domain as an extra dimension. The tubes are matched along the spatial domain and completion is performed on the stacked tubes. However, tube matching is a batch method and is not applicable to {streaming} tensor completion since the information in the temporal domain is not available. Further, as Fig.\ref{fig:insight} depicts, tube matching ignores the change in motion along the temporal dimension. In this work, we develop a novel efficient strategy for patch matching. We perform patch matching on dilated tensor data by propagating the observed pixels to unobserved neighboring pixels. Then, a new efficient coarse-scale patch matching strategy is proposed to efficiently match the patches with missing data. 

To accurately complete the missing entries of a streaming tensor formed from tracked patches, we develop a new {streaming} tensor completion algorithm using the tensor ring model. This model exploits a more compact expression for high-order tensors, and has shown superior performance in tensor completion, especially with highly incomplete data \cite{yu2020low}. Compared to CP-based {streaming} tensor completion methods \cite{mardani2015subspace}, which are only designed for $3$-way tensors, the proposed {streaming} tensor ring completion algorithm can work with higher-order tensors required by the proposed framework. Compared with TOUCAN \cite{gilman2020online} and STC \cite{xie2018accurate}, the proposed method utilizes the more compact latent core tensors to track the tensor subspace. We propose a fast {streaming} completion algorithm for the tensor ring model using a least-squares and one-step scaled steepest descent method, which can efficiently update the latent core tensors and complete the streaming tensor with high accuracy. 

The main contributions of the paper are summarized as follows:

1. We develop a novel patch tracking-based {streaming} tensor ring completion framework for visual data recovery. An efficient coarse-scale patch matching strategy is proposed for patch tracking with missing data. To the best of our knowledge, this is the first patch-based framework { that incorporates a streaming tensor completion method}. 

2. We propose a new {streaming} tensor completion algorithm using the tensor ring model. The algorithm utilizes least-squares and one-step scaled steepest descent, and can efficiently update the latent core tensors and complete the tracked patch tensor.

3. We conduct extensive experiments that demonstrate the superior performance of the proposed method over state-of-the-art {batch} and {streaming} tensor completion algorithms. 

The paper is organized as follows. In Section \ref{sec:bkgnd}, we briefly introduce our notation and provide some preliminary background on the tensor ring and its properties. In Section \ref{sec:nlptronline}, we propose the new patch-based {streaming} tensor ring completion framework. In Section \ref{sec:onlinetrc}, we develop the {streaming} tensor ring completion algorithm. {The complexity analysis is given in Section \ref{sec:complexity}.} Experimental results are presented in Section \ref{sec:results} to demonstrate the completion performance. Finally, the conclusion is given in Section \ref{sec:conclusion}.

\section{Preliminaries}
\label{sec:bkgnd}
Uppercase script letters are used to denote tensors (e.g., ${\cal X}$), boldface uppercase letters to denote matrices (e.g., ${\mathbf{X}}$) and boldface lowercase letters to denote vectors (e.g., ${\mathbf{x}}$). An $N$-order tensor is defined as ${\cal X}\in{\mathbb{R}}^{I_1\times \ldots \times I_N}$, where $I_i,i\in[N]:=\{1,\ldots, N\}$ is the dimension of the $i$-th way of the tensor, and ${\cal{X}}_{i_1\ldots i_N}$ denotes the $(i_1,i_2,...,i_N)$-th entry of tensor $\mathbf{\cal{{X}}}$. The vector ${\mathbf{x}}\in\mathbb{R}^{\prod_{k=1}^N I_k}$ denotes the vectorization of tensor $\mathcal{X}$. For a $3$-way order tensor (i.e., $N=3$), the notation ${{\cal{X}}}(:,:,i),{{\cal{X}}}(:,i,:),{{\cal{X}}}(i,:,:)$ denotes the frontal, lateral, horizontal slices of $\mathbf{\cal{X}}$, respectively. The Frobenius norm of a tensor ${\cal X}$ is defined as $\|{{\cal{X}}}\|_F=\sqrt{\sum_{i_1\ldots i_N}|{{\cal{X}}_{i_1\ldots i_N}}|^2}$. The product $\mathcal{A}\circ\mathcal{B}$ denotes the Hadamard (element-wise) product of two tensors $\mathcal{A}$ and $\mathcal{B}$. Next, we introduce some definitions and theorems for tensor ring used throughout the paper. 

\begin{mydef}
[TR Decomposition \cite{zhao2016tensor}] In TR decomposition, a high-order tensor $\mathcal{X} \in \mathbb{R}^{I_{1} \times \cdots \times I_{N}}$ can be represented using a sequence of circularly contracted 3-order core tensors $\mathcal{Z}_k\in\mathbb{R}^{r_k\times I_k\times r_{k+1}},k=1,\ldots,N, r_{N+1}=r_1$. Specifically, the element-wise relation of tensor $\mathcal{X}$ and its TR core tensors $\{\mathcal{Z}_k\}_{k=1}^N$ is defined as
\begin{equation}
\mathcal{X}_{i_{1} \ldots i_{N}}=\operatorname{Tr}\left(\prod_{k=1}^N\mathcal{Z}_k(:,i_k,:)\right)\:,
\label{eq:TR_decomp}
\end{equation} 
where $\operatorname{Tr}(\cdot)$ is the matrix trace operator. The relation of tensor $\mathcal{X}$ and its TR decomposition core tensors $\{\mathcal{Z}_k\}_{k=1}^N$ is written as $\mathcal{X}=\Re\left(\mathcal{Z}_{1}, \mathcal{Z}_{2}, \ldots, \mathcal{Z}_{N}\right)$, where $\Re$ is the function defined through \eqref{eq:TR_decomp}, and  $[r_1,\ldots,r_N]$ is called TR rank.
\label{def:TR_dec}
\end{mydef}

\begin{mydef}
[Mode-$k$ unfolding \cite{yu2019tensor,yu2020low}]
The mode-k unfolding matrix of tensor $\mathcal{X}$ is denoted by $\mathbf{X}_{[k]}$ of size $n_{k} \times \prod_{j \neq k} n_{j}$ with its elements defined by
\[
{X}_{[k]}\left(i_{k}, \overline{i_{k+1}, \ldots, i_{N} i_{1}, \ldots, i_{k-1}}\right)=\mathcal{X}\left(i_{1}, i_{2}, \ldots, i_{N}\right)\:.
\]
Another classical mode-$k$ unfolding matrix of $\mathcal{X}$ 
is denoted by $\mathbf{X}_{(k)}$ of size $n_{k} \times \prod_{j \neq k} n_{j}$ and defined by
\[
{X}_{(k)}\left(i_{k}, \overline{i_{1}, \ldots, i_{k-1} i_{k+1}, \ldots, i_{N}}\right)=\mathcal{X}\left(i_{1}, i_{2}, \ldots, i_{N}\right)\:.
\]

\end{mydef}

\begin{mytheo}[\cite{zhao2016tensor}]
Given a TR decomposition $\mathcal{X}=\Re\left(\mathcal{Z}_{1}, \ldots, \mathcal{Z}_{N}\right)$, its mode-$k$ unfolding matrix $\mathbf{X}_{[k]}$ can be written as
$$
\mathbf{X}_{[k]}=\mathbf{Z}_{k(2)}\left(\mathbf{Z}_{[2]}^{\neq k}\right)^{T}
$$
where $\mathbf{Z}^{\neq k}\in\mathbb{R}^{r_{k+1} \times \prod_{j=1, j \neq k}^{N} I_{j} \times r_{k}}$ is a subchain obtained by merging all cores except $\mathcal{Z}^{k}$, whose slice matrices are defined by 
$$
\mathbf{Z}^{\neq k}\!\left(\overline{i_{k+1} \cdots i_{N} i_{1} \ldots i_{k-1}}\right)\!=\!\!\prod_{j=k+1}^{N}\! \!\mathcal{Z}_{j}\left(:,i_{j},:\right) \prod_{j=1}^{k-1} \mathcal{Z}_{j}\left(:,i_{j},:\right)\:.
$$
\label{thm1}
\end{mytheo}

\section{Patch tracking-based {streaming} tensor ring completion framework}
\label{sec:nlptronline}
 
\begin{figure*}[htbp]
\centering
\includegraphics[width=1\linewidth]{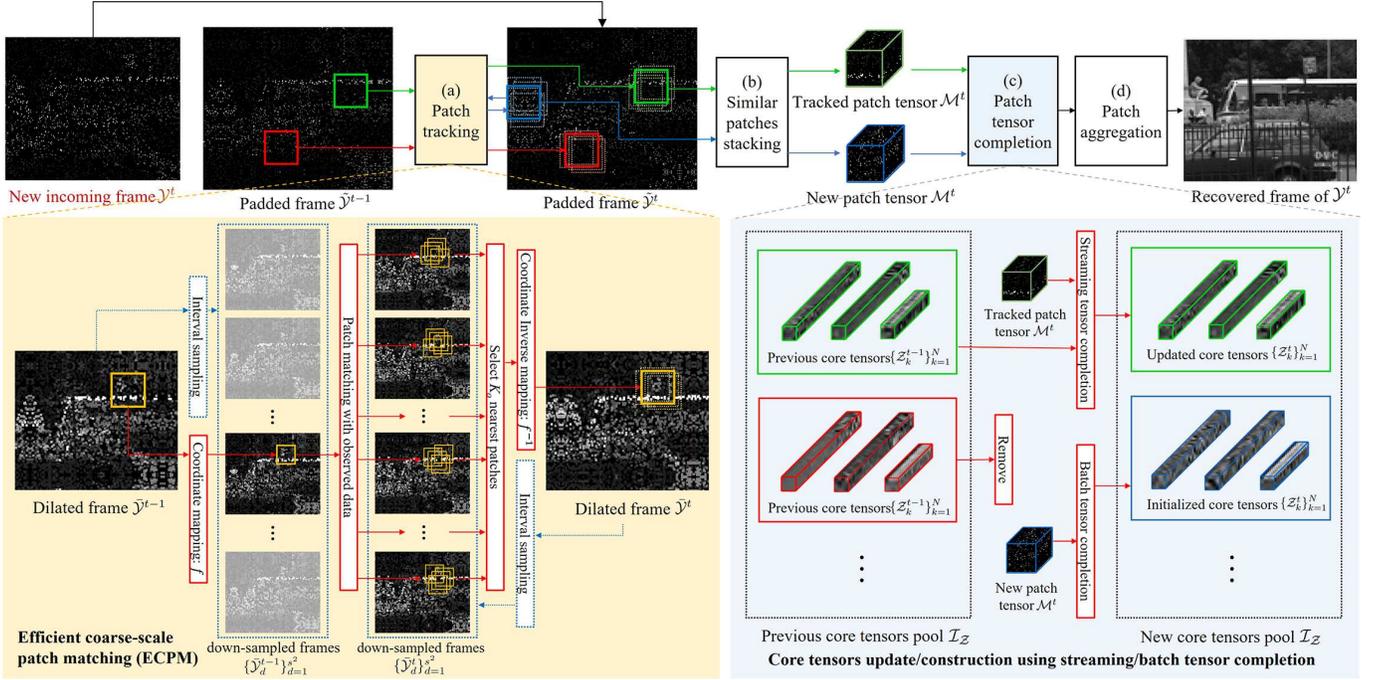}
\caption{{Illustration of the proposed patch tracking-based streaming tensor ring completion framework. Given a newly incoming frame $\mathcal{Y}^t$, (a) an efficient patch tracking method is firstly applied to track the patches from frame $\mathcal{Y}^{t-1}$. Specifically, for tracked patches of $\mathcal{Y}^{t-1}$ (shown as green and red solid rectangles), $K_o$ most similar patches in $\mathcal{Y}^{t}$ are matched using the proposed ECPM method (the first similar patch with smallest distance is shown as a solid rectangle, while the remaining $K_o-1$ patches are shown as dotted rectangles). The blue solid rectangle designates a new patch for an uncovered region in $\mathcal{Y}^{t}$, and the dotted blue rectangles around it denote its $K_b-1$ most similar patches. (b) Patch tensors are created by stacking $K_o$ (or $K_b$) nearest similar green (or blue) patches. (c) Core tensors pool is updated using the corresponding patch tensors, and subsequently patch tensors are completed using the updated core tensors pool. Here, we assume that the red tracked patch in $\mathcal{Y}^{t}$ is mistracked or has large overlap with other tracked patches, and should be removed from the tracking process. (d) Finally, the recovered new frame is obtained by aggregating the completed patch tensors.}}
\label{fig:PTSTRC}
\end{figure*}

In this section, we propose our patch tracking-based {streaming} tensor ring completion (PTSTRC) framework for streaming visual data recovery, which is illustrated in Fig.~\ref{fig:PTSTRC}. First, we start with a high-level description of the proposed framework, followed by a detailed description of the different procedures used. 

Assume we have a partially observed video sequence $\{\mathcal{Y}^{i}\}_{i=1}^T$, $\mathcal{Y}^{i}\in\mathbb{R}^{I_1\times I_2\times n}$, with corresponding observed pixel index sets $\{\Omega^i\}_{i=1}^T$. The number of frames $T$ can be sufficiently large such that the video is a streaming video. For gray-scale and color frames, the value of $n$ is $1$ and $3$, respectively. The video frames are given sequentially, i.e., the video frame $\mathcal{Y}^{t}$ arrives at time $t$. To match the patches at the corners and boundaries of the frames, the arriving frame $\mathcal{Y}^{t}$ is first padded by mirroring $b$ pixels at all boundaries and corners, resulting in a padded frame $\tilde{\mathcal{Y}}^{t}$ with frame size ${(I_1+2b)\times (I_2+2b)\times n}$ and corresponding observation index sets $\tilde{\Omega}^t$.

For the initial video frame ${\mathcal{Y}}^{1}$, after padding it to $\tilde{\mathcal{Y}}^{1}$, we divide it into overlapping patches of size $m\times m\times n$ with a number of overlap pixels $o$, and set each of these patches as a 'tracked patch' of $\tilde{\mathcal{Y}}^{1}$. For each tracked patch, we find the $K_b-1$ most similar patches within a search window of size $l\times l$ centered at the location of the tracked patch. Then, a patch tensor of size $m\times m \times n \times K_b$ is created by stacking the tracked patch and its $K_b-1$ similar patches.

At time $t>1$, for each tracked patch of $\tilde{\mathcal{Y}}^{t-1}$, a patch matching method is used to find the $K_o$ most similar patches in $\tilde{\mathcal{Y}}^{t}$ within a search window of size $l\times l$. Meanwhile, the first similar patch (i.e., the one with the smallest distance) is designated as a tracked patch of $\tilde{\mathcal{Y}}^t$ {and will be utilized for patch tracking of the next incoming frame.} Then, a tracked patch tensor of size $m\times m\times n \times K_o$ is formed by stacking the $K_o$ most similar patches. Further, patch refinement is invoked to prune the {mistracked or} highly overlapping tracked patches, {as well as create new patches for uncovered regions.} 

After obtaining all the patch tensors required for completion of $\tilde{\mathcal{Y}}^{t}$, the relevant core tensors for each patch tensor are updated (or created) using a {streaming} (or {batch}) tensor ring completion algorithm. Then, the core tensors are stored in a tensor pool for completion of the next frame. Finally, the frame at time $t$ is recovered as $\mathcal{P}^{t}\circ{\mathcal{Y}}^{t}+(1-\mathcal{P}^{t})\circ\hat{\mathcal{Y}}^{t}$, {where the observation mask $\mathcal{P}^{t}$ is defined by}
\begin{equation}
{\cal{P}}_{i_1,i_2,i_3}^{t}=\left\{\begin{array}{cl}
1, & \text{if }(i_1,i_2,i_3) \in \Omega^{t} \\
0, & \text {otherwise}
\end{array}\right.
\end{equation}
and $\hat{\mathcal{Y}}^{t}$ is obtained by aggregating all the recovered patches belonging to $\tilde{\mathcal{Y}}^{t}$ and removing padded border pixels. 

As shown in Fig.~\ref{fig:PTSTRC}, the framework contains two main procedures, namely, patch tracking and patch tensor completion. Next, we discuss the different components underlying our patch tracking procedure. {streaming} tensor completion will be discussed in the next section.

\subsection{Image dilation for patch tracking with missing data}

{Compared with traditional patch tracking methods\cite{dabov2007image,ji2010robust,maggioni2012nonlocal,wen2017joint} where the data is fully observed, patch tracking with a portion of observed data is more challenging,} especially when the number of observed pixels is relatively small. For example, for a patch of size $20 \times 20$ and observation ratio $20\%$, the number of observed pixels common between two patches will be about $16$ in average. In this case, the matching accuracy may degrade due to lack of enough common information. In this paper, we address this problem using a simple but efficient image dilation method. Specifically, the $(i,j)$-th entry of the dilated frame $\bar{\mathcal{Y}}$ for the $n$-th channel is given by
\begin{equation}
\bar{\mathcal{Y}}_{i,j,n}=
\left\{  
\begin{array}{cc}
\max\limits_{(p,q)\in\mathcal{N}_{i,j}, (p,q,n)\in\tilde{\Omega}}\tilde{\mathcal{Y}}_{p,q}, & \mathcal{N}_{i,j}\cap\tilde{\Omega}\neq\varnothing\\
\rm{NaN} , & \text {otherwise}  
\end{array}  
\right.  
\end{equation}
where $\mathcal{N}_{i,j}$ is the index set formed by $9$ neighbors of index $(i,j)$ (including itself), and ${\rm{NaN}}$ denotes that the corresponding entry is missing.
  
There are two advantages in using dilation. First, the observed pixels are propagated to the unobserved neighboring pixels, so the observation percentage is greatly improved. Second, the maximization operation among adjacent observed pixels brings larger positional margin to matching, which will be useful for the next coarse-scale matching step. We remark that using a dilated image for matching may yield patches that are within a small offset from the true matching positions. However, since it is a small offset, it does not affect the completion performance significantly, which will be verified in the experiments.

\subsection{Efficient coarse-scale patch matching}
Another bottleneck for the patch-based method is the computational efficiency. Searching similar patches could take a long time if the size of the search window is large. On the other hand, a small search window size may decrease the matching performance. To improve the efficiency of patch matching, one could make use of a hierarchical or multi-resolution search method \cite{huang2006survey,hussain2019survey}, which searches for similar patches on different coarse to fine scales. However, due to the existence of missing entries, {the traditional interpolation-based downscale method \cite{je2013optimized} cannot be applied to the partially observed frames}, which precludes direct use of the hierarchical search method for our matching task. 

To improve the matching efficiency, in this work we propose a new {efficient} coarse-scale patch matching (ECPM) method, illustrated in the bottom-left of Fig.~\ref{fig:PTSTRC}. At time $t$, given a sampling interval parameter $s$, both dilated new frame $\bar{\mathcal{Y}}^{t}$ and previous frame $\bar{\mathcal{Y}}^{t-1}$ are down-sampled by selecting pixels with interval $s$. Then, the corresponding $s^2$ down-sampled frames $\{\bar{\mathcal{Y}}^{t}_{d}\}_{d=1}^{s^2}$ and $\{{\bar{\mathcal{Y}}^{t-1}_{d}}\}_{d=1}^{s^2}$ of size $(I_1+2b)/s\times(I_2+2b)/s$ are obtained (here we assume that $b$ is such that $I_1+2b$ and $I_2+2b$ are divisible by $s$). Patch matching is performed on the down-sampled frames and finally the locations of the matched patches in $\bar{\mathcal{Y}}^{t}$ are obtained. The detailed steps of the proposed ECPM method are described next. To simplify the expressions, we use the coordinate of the top left pixel of the patch to represent the patch location.

\subsubsection{Build pixel-wise coordinate mapping between original frame and down-sampled frames} The mapping function $f:\mathbb{R}^2\rightarrow\mathbb{R}^3$ is defined as $f: (x,y)\mapsto(x',y',c)$ where 
\begin{equation}
\begin{aligned}
&x'=\left\lfloor\frac{x-1}{s}\right\rfloor+1, ~y'=\left\lfloor\frac{y-1}{s}\right\rfloor+1,\\
&c={\rm{mod}}(x-1,s)+{\rm{mod}}(y-1,s)\times s+1
\end{aligned}
\label{eq:mapping}
\end{equation}
with $c\in[1,s^2]$, such that the pixel location $(x,y)$ in the original image is mapped to location $(x',y')$ in the $c$-th down-sampled frame. 

\subsubsection{Perform patch matching on down-sampled frames} Given the tracked patch located at $(x_0,y_0)$ in the previous frame $\bar{\mathcal{Y}}^{t-1}$, we obtain its location $(x_0',y_0')$ in the $c_0$-th down-sampled frame $\bar{\mathcal{Y}}^{t-1}_{c_0}$ using \eqref{eq:mapping}. Then, the patch with size $\lceil m/s \rceil\times\lceil m/s \rceil\times n$ at $(x_0',y_0')$ in $\bar{\mathcal{Y}}^{t-1}_{c_0}$ is utilized as the down-sampled tracked patch, and patch matching is performed across all the $s^2$ down-sampled frames $\{\bar{\mathcal{Y}}^{t}_{d}\}_{d=1}^{s^2}$ of $\bar{\mathcal{Y}}^{t}$ within a search region of size $\lceil l/s \rceil\times\lceil l/s \rceil$ centered at $(x_0',y_0')$. {In particular, given two patches $\mathcal{S}_1$ and $\mathcal{S}_2$ with corresponding observation masks $\mathcal{P}_1$ and $\mathcal{P}_2$, the distance (dissimilarity) is measured as
\begin{equation}
\begin{aligned}
d(\mathcal{S}_1,\mathcal{S}_2)=\frac{\|\mathcal{P}_1\circ\mathcal{P}_2\circ(\mathcal{S}_1-\mathcal{S}_2)\|^2_F}{\|\mathcal{P}_1\circ\mathcal{P}_2\|_0}
\end{aligned}
\end{equation}
where the $\ell_0$-norm $\|\cdot\|_0$ counts the total number of non-zero elements. {A larger distance $d$ indicates smaller similarity.}} The $K_o$ most similar patches of the tracked patch in $\{\bar{\mathcal{Y}}^{t}_{d}\}_{d=1}^{s^2}$ are obtained with locations $\{(x'_i,y'_i,c_i)\}_{i=1}^{K_o}$.

\subsubsection{Map the matched patches to original location} The locations of the $K_o$ most similar patches in the new frame $\bar{\mathcal{Y}}^{t}$ (and $\tilde{\mathcal{Y}}^{t}$) can be computed using the following inverse mapping $f^{-1}: (x',y',c)\mapsto(x,y)$ where
\begin{equation}
\begin{aligned}
&x=(x'-1)s+1+{\rm{mod}}(c-1,s)\\
&y=(y'-1)s+1+\left\lfloor\frac{c-1}{s}\right\rfloor\:,
\end{aligned}
\end{equation}

Note that patch matching on each down-sampled frame is independent, so parallelization could be used to further improve efficiency. Using the above ECPM method, for each tracked patch in $\tilde{\mathcal{Y}}^{t-1}$, the $K_o$ most similar patches in the new frame $\tilde{\mathcal{Y}}^{t}$ are obtained and stacked to construct a patch tensor of size $m \times m \times n \times K_o$. Specifically, patch matching can be applied within the frame $\tilde{\mathcal{Y}}^{t-1}$ itself by setting the new frame as $\tilde{\mathcal{Y}}^{t-1}$. We remark that the coarse-scale matching may also introduce a small offset between the tracked position and the true matching positions. However, as verified in the experiments, such a method achieves significant speedup with only slight degradation in recovery performance.

\subsection{Patch refinement}
During the patch tracking process, there are three types of patches to be processed at a given time $t$, marked in green, red and blue in Fig.~\ref{fig:PTSTRC}. {In particular, correctly tracked patches (solid green rectangle in $\tilde{\mathcal{Y}}^{t}$) will be used for completion of $\tilde{\mathcal{Y}}^{t}$ and tracking of the next frame $\tilde{\mathcal{Y}}^{t+1}$. Some mistracked or highly overlapping patches (solid red rectangle in $\tilde{\mathcal{Y}}^{t}$) should be removed from the tracking procedure. Further, new patches (solid blue rectangle in $\tilde{\mathcal{Y}}^{t}$) should be created to cover regions that are not covered by the tracked patches. The specific operations for the latter two types of patches are detailed next.}

\subsubsection{Pruning mistracked or highly overlapping patches}

{Long video data may contain scene variations or abrupt changes between adjacent frames. In this case, patch tracking may result in tracking incorrect patches with low similarity. Therefore, such mistracked patches should be removed from the tracking procedure. To this end, we set a threshold $\tau_f$ to decide if the patch tracking is `success' or `failure'. {For a tracked patch of $\tilde{\mathcal{Y}}^{t}$, if its distance to the corresponding tracked patch in $\tilde{\mathcal{Y}}^{t-1}$ is larger than $\tau_f$,} this patch tracking process is deemed a failure and the tracked patch is removed. The corresponding core tensors in the core tensors pool are also removed.}

It can be also observed that tracked patches may have considerable overlap after several tracking iterations. Continuing to track all such patches will result in no or marginal improvement in the completion performance at the expense of extra computational and storage cost. Thus, we stop tracking of these patches by detecting their degree of overlap. Specifically, the overlap degree of a tracked patch is obtained by firstly counting the number of times each pixel is shared with other tracked patches, and then picking the minimum number. If the overlap degree is larger than a predefined threshold $\tau_c$, the corresponding tracked patch is removed. 

\subsubsection{Generating new patches for uncovered regions}
{Some regions in the new frame $\tilde{\mathcal{Y}}^t$ may not be covered by tracked patches from $\tilde{\mathcal{Y}}^{t-1}$ due to the emergence of new scenes or the removal of mistracked patches.} We create new overlapping patches of size $m \times m\times n$ with a number of overlap pixels $o$ to cover these regions. These new patches are added to the tracked patches of $\tilde{\mathcal{Y}}^t$. {Then, for each new patch, the ECPM method is applied to find the $K_b-1$ most similar patches in $\tilde{\mathcal{Y}}^{t}$ around it within a search window of size $l \times l$, and a new patch tensor of size $m\times m \times n \times K_b$ is created by stacking the new patch and its $K_b-1$ similar patches.} Finally, a batch tensor completion algorithm is applied to the new patch tensors and the obtained core tensors are added to the core tensors pool.

The entire process of the proposed PTSTRC method at time $t$ is summarized in Algorithm \ref{alg:PTSTRC}. It should also be noted that, for tracked patch set $\mathcal{I}_{\mathcal{S}}$, only the patch location information need to be stored. Further, one can recover $\mathcal{Y}^1$ at $t=1$ by setting $\mathcal{I}_{\mathcal{S}}=\mathcal{I}_{\mathcal{Z}}=\varnothing$, i.e., treating $\mathcal{Y}^1$ as a frame with an entirely uncovered region.

{According to Algorithm \ref{alg:PTSTRC}, to complete the new frame $\mathcal{Y}^t$, in practice one should store the previous frame $\mathcal{Y}^{t-1}$ with the corresponding observed pixel index set $\Omega^{t-1}$, the coordinates of tracked patches of $\tilde{\mathcal{Y}}^{t-1}$, and the core tensors corresponding to the tracked patches. Taking advantage of the TR decomposition model, the past information is fused into the core tensors so the algorithm can run in a streaming fashion without storage of frames before $t-1$.}

\begin{algorithm}
\caption{Patch tracking-based {streaming} tensor ring completion (PTSTRC)}
\begin{algorithmic}[1]
 \REQUIRE New incoming frame $\mathcal{Y}^{t}$ with corresponding observed pixel set $\Omega^t$, previous frame $\mathcal{Y}^{t-1}$ with corresponding $\Omega^{t-1}$, previous tracked patch set $\mathcal{I}_{\mathcal{S}}$ with corresponding core tensors pool (set) $\mathcal{I}_{\mathcal{Z}}$, patch matching parameters $b$, $m$, $o$, $s$, $l$, $K_b$ and $K_o$, {thresholds $\tau_f$ and $\tau_c$}. 
 \STATE Pad the frames and get $\tilde{\mathcal{Y}}^{t-1}$ and $\tilde{\mathcal{Y}}^{t}$.
 \FOR {each patch $\mathcal{S}_i$ in $\mathcal{I}_{\mathcal{S}}$}
 \STATE Search $K_o$ nearest patches within search range $l\times l$ in $\tilde{\mathcal{Y}}^t$. 
 {
 \IF {Smallest distance $>\tau_f$ or overlap degree $>\tau_c$}
 \STATE Remove $\mathcal{S}_i$ from $\mathcal{I}_{\mathcal{S}}$, remove corresponding core tensors from $\mathcal{I}_{\mathcal{Z}}$.
 \ELSE
 \STATE Replace $\mathcal{S}_i$ in $\mathcal{I}_{\mathcal{S}}$ with its first similar patch in $\tilde{\mathcal{Y}}^t$. Construct patch tensor $\mathcal{M}_i^t$ using the obtained $K_o$ most similar patches. Update core tensors $\{\mathcal{Z}_{i,k}\}_{k=1}^{N}$ related to $\mathcal{S}_i$ using $\mathcal{M}_i^t$. 
 \ENDIF }
 \ENDFOR
 \STATE Create new tracked patches set $\mathcal{I}'_{\mathcal{S}}$ to cover the uncovered region in $\tilde{\mathcal{Y}}_t$ according to $\mathcal{I}_{\mathcal{S}}$. 
 \FOR {each patch $\mathcal{S}_i$ in $\mathcal{I}'_{\mathcal{S}}$}
 \STATE Search $K_b$ nearest patches within search range $l\times l$ in $\tilde{\mathcal{Y}}_t$. Add $\mathcal{S}_i$ to $\mathcal{I}_{\mathcal{S}}$. Construct patch tensor $\mathcal{M}_i^t$ using obtained $K_b$ nearest patches. 
 \STATE Compute core tensors $\{\mathcal{Z}_{i,k}\}_{k=1}^{N}$ related to $\mathcal{S}_i$ and add them to $\mathcal{I}_{\mathcal{Z}}$.
 \ENDFOR
 \STATE Obtain $\hat{\mathcal{Y}}^{t}$ by aggregating patches according to $\mathcal{I}_{\mathcal{Z}}$ and remove padded border pixels.
 \STATE Compute recovered frame as $\mathcal{P}^{t}\circ{\mathcal{Y}}^{t}+(1-\mathcal{P}^{t})\circ\hat{\mathcal{Y}}^{t}$.
\ENSURE Recovered frame of $\mathcal{Y}^t$, new tracked patch set $\mathcal{I}_{\mathcal{S}}$ with corresponding core tensors pool (set) $\mathcal{I}_{\mathcal{Z}}$.
\end{algorithmic}
\label{alg:PTSTRC}
\end{algorithm}

{\begin{remark}
Algorithm \ref{alg:PTSTRC} can be simplified to a patch matching (PM)-based image completion algorithm, where only one incoming frame (image) is given and needs to be completed. 
In this case, step 2 to step 9 are removed (no tracking needed). For step 13, the completion is no longer limited to tensor-decomposition-based algorithms since there is no need to construct or update core tensors. For example, $\mathcal{M}_i^t$ can be completed using nuclear-norm-based algorithms such as tensor ring nuclear norm (TRNN)\cite{yu2019tensor} and tensor train nuclear norm (TTNN)\cite{bengua2017efficient}. However, nuclear-norm-based algorithms are not applicable to streaming tensor completion with sequential incoming frames because they cannot construct and update core tensors.
\end{remark}
}

\section{{streaming} tensor ring completion algorithm}
\label{sec:onlinetrc}

In this section, we develop an efficient {streaming} tensor completion algorithm for patch tensor completion. We adopt a tensor ring model to formulate the {streaming} tensor completion, which is verified to achieve better completion performance compared with the CP and Tucker model. To date, there is no {streaming} version for tensor ring completion. {Further, we develop a new batch tensor ring completion algorithm to complete the new patch tensors and create the corresponding core tensors.}

\subsection{Tensor ring completion}
First, let us revisit traditional tensor ring completion. Given a tensor $\mathcal{M}\in\mathbb{R}^{I_1\times I_2\times \ldots \times I_N}$ and its observed index set $\Omega$, the recovered full tensor $\mathcal{X}$ can be obtained by solving the following minimization problem
\begin{equation}
\min_{\mathcal{X}}\operatorname{rank}_t(\mathcal{X})\text{, s.t. } \mathcal{P}\circ\mathcal{X}=\mathcal{P}\circ\mathcal{M}
\label{eq:tc_st}
\end{equation}
where $\operatorname{rank}_t(\mathcal{X})$ is a tensor rank of $\mathcal{X}$. 

According to Definition \ref{def:TR_dec}, the cost function for the tensor ring completion algorithm can be formulated as
\begin{equation}
\min _{\mathcal{Z}_{1}, \ldots, \mathcal{Z}_{N}}\left\|\mathcal{P}\circ(\mathcal{M}-\Re(\mathcal{Z}_{1}, \ldots, \mathcal{Z}_{N}))\right\|_{F}^2\:.
\label{eq:trals}
\end{equation}
The above problem can be solved using alternating least squares (ALS) \cite{yuan2018higher}. Specifically, at each iteration, the core tensor $\mathcal{Z}_k$ is updated by fixing all core tensors except $\mathcal{Z}_k$, and based on Theorem \ref{def:TR_dec}, the following problem is solved
\begin{equation}
\min_{\mathcal{Z}_{k}}\left\|\mathbf{P}_{[k]}\circ(\mathbf{M}_{[k]}-\mathbf{Z}_{k(2)}\left(\mathbf{Z}_{[2]}^{\neq k}\right)^{T})\right\|_{F}^2\:.
\end{equation}
The final estimated tensor $\mathcal{X}$ can be calculated by $\mathcal{X}=\Re(\mathcal{Z}_{1}, \ldots, \mathcal{Z}_{N})$.

\subsection{{Streaming} tensor ring completion}
Specific to the proposed patch-based {streaming} tensor ring completion framework, for a tracked patch in $\tilde{\mathcal{Y}}^{t-1}$, $K_o$ nearest patches are found in $\tilde{\mathcal{Y}}^{t}$ and formed as a patch tensor  $\mathcal{M}^t\in\mathbb{R}^{I_1 \times \ldots \times I_{N-1}\times K_o}$ with corresponding mask tensor $\mathcal{P}^t$. Further, it can be observed from \eqref{eq:trals} that only the last core tensor $\mathcal{Z}_{N}$ incorporates the temporal information of $\mathcal{M}$ (i.e., related to the temporal dimension). Thus, denoting by $\mathcal{V}^i\in\mathbb{R}^{r_N\times K_o \times r_1}$ the subtensor of $\mathcal{Z}_N$ related to $\mathcal{M}^i$, the tensor ring completion problem at time $t$ using \eqref{eq:trals} can be rewritten as 
\begin{equation}
\begin{aligned}
\!\min_{\mathcal{Z}_{1}^t, \ldots, \mathcal{V}^{i}}&\!\sum_{i=1}^{t}\!\left\|\mathcal{P}^i\!\circ\!\left(\mathcal{M}^i \!-\!\Re(\mathcal{Z}_{1}^t, \!\ldots\!, \mathcal{Z}_{N-1}^t,\mathcal{V}^i)\right)\right\|_{F}^2 \!+\!\gamma\!\sum_{i=1}^{t}\!\|\mathcal{V}^i\|_F^2
\label{eq:online1}
\end{aligned}
\end{equation}
where we also included a regularization term to constrain the norm of each subtensor $\mathcal{V}^i$. $\gamma$ is the regularization parameter.

{For batch tensor ring completion \eqref{eq:online1} (or \eqref{eq:trals}), all tensors $\{\mathcal{M}^i\}_{i=1}^t$ and $\{\mathcal{P}^i\}_{i=1}^t$ must be accessible simultaneously for the completion process. However, as described in Algorithm 1, only the latest tensor $\mathcal{M}^t$ with the corresponding $\mathcal{P}^t$ and the previous core tensors $\{\mathcal{Z}_k^{t-1}\}_{k=1}^{N}$ are available at time $t$, which precludes the use of the current batch-based tensor ring completion methods. In this work, we develop a new streaming tensor ring completion algorithm to solve \eqref{eq:online1}. The proposed method uses least-squares to get an estimate of $\mathcal{V}^t$ using $\{\mathcal{Z}_k^{t-1}\}_{k=1}^{N-1}$, followed by an approximation of $\{\mathcal{Z}_k^{t}\}_{k=1}^{N-1}$ using a scaled steepest descent (SSD) method.}

\subsubsection{Computation of $\mathcal{V}^t$}

Suppose that at time $t$ we have the tensor $\mathcal{M}^t$ and the previous estimated core tensors $\{\mathcal{Z}_k^{t-1}\}_{k=1}^{N-1}$. The optimization problem in terms of $\mathcal{V}^t$ in \eqref{eq:online1} becomes
\begin{equation}
\begin{aligned}
{\mathcal{V}^t=\arg\min_{\mathcal{V}}\left\|\mathbf{P}^t_{[N]}\!\circ\!(\mathbf{M}^t_{[N]} \!-\!\mathbf{V}_{(2)}(\mathbf{U}^{t-1})^T)\right\|_{F}^2+\gamma\|\mathcal{V}\|_F^2}
\label{eq:onlines}
\end{aligned}
\end{equation}
where $\mathbf{U}^{t-1}=(\mathbf{Z}^{t-1})_{[2]}^{\neq N}$. Defining $\mathbf{v}_k\in\mathbb{R}^{r_Nr_1}$ as the vectorized form of the $k$-th lateral slice of $\mathcal{V}^t$, the solution of \eqref{eq:onlines} can be obtained by solving the following $K_o$ sub-problems:
\begin{equation}
\mathbf{v}_k\!=\!\arg\!\min_{\mathbf{v}\in\mathbb{R}^{r_Nr_1}}\!\!\left\|\mathbf{p}_k\!\circ\!(\mathbf{m}_k\!-\!\mathbf{U}^{t-1}\mathbf{v})\right\|_{2}^2+\gamma\|\mathbf{v}\|_2^2,k=1,\ldots,K_o
\label{eq:online3}
\end{equation}
where $\mathbf{p}_{k}$ and $\mathbf{m}_k$ are the $k$-th {rows} of {$\mathbf{P}^t_{[N]}$} and {$\mathbf{M}^t_{[N]}$}, respectively. Further, \eqref{eq:online3} can be simplified as
\begin{equation}
\label{subproblem}
\mathbf{v}_k=\arg\min_{\mathbf{v}\in\mathbb{R}^{r_Nr_1}}{\left\| [\mathbf{m}_{k}]_{\mathcal{I}_k}-[\mathbf{U}^{t-1}]_{\mathcal{I}_k}\mathbf{v}\right\|_{2}^2}+\gamma\|\mathbf{v}\|_F^2
\end{equation}
where $\mathcal{I}_k$ denotes the index set of the non-zero entries of $\mathbf{p}_{k}$, and $[\mathbf{m}_{k}]_{\mathcal{I}_k}\in\mathbb{R}^{|\mathcal{I}_k|}$ denotes the vector of  elements in $\mathbf{m}_k$ indexed by set $\mathcal{I}_k$ of cardinality $|\mathcal{I}_k|$. The matrix $[\mathbf{U}^{t-1}]_{\mathcal{I}_k}\in\mathbb{R}^{|\mathcal{I}_k|\times r_Nr_1}$ is formed from the rows of $\mathbf{U}^{t-1}$ with row index $\mathcal{I}_k$. 
Eq. (\ref{subproblem}) has a closed-form solution
\begin{equation}
\mathbf{v}_k=\left([\mathbf{U}^{t-1}]_{\mathcal{I}_k}^T[\mathbf{U}^{t-1}]_{\mathcal{I}_k}+\gamma\mathbf{I}\right)^{-1}[\mathbf{U}^{t-1}]_{\mathcal{I}_k}^T[\mathbf{m}_{k}]_{\mathcal{I}_k}\:.
\label{eq:online5}
\end{equation}

\subsubsection{Update $\mathcal{Z}_{1}^t, \ldots, \mathcal{Z}_{N-1}^t$}

Traditional {streaming} tensor completion methods usually apply stochastic gradient descent (SGD) or recursive least squares (RLS) to update the relative tensors. However, as shown in \cite{kasai2016online}, directly applying SGD suffers from a low convergence rate, while the efficiency of RLS is limited due to its high computational complexity. In this work, we apply scaled steepest descent (SSD) method \cite{tanner2016low} to solve $\mathcal{Z}_{1}^t, \ldots, \mathcal{Z}_{N-1}^t$. 

By only considering the data available at time $t$, {we have $\mathcal{Z}_N^t=\mathcal{V}^t$} and the optimization problem in \eqref{eq:online1} at time $t$ reduces to
\begin{equation}
\begin{aligned}
\!\min_{\mathcal{Z}_{1}^t, \ldots, \mathcal{Z}_{N-1}^t}\!\mathcal{L}:=\!\left\|\mathcal{P}^t\!\circ\!\left(\mathcal{M}^t \!-\!\Re(\mathcal{Z}_{1}^t, \ldots, \mathcal{Z}_{N-1}^t,\mathcal{Z}_N^t)\right)\right\|_{F}^2 
\label{eq:online11}
\end{aligned}
\end{equation}
Traditional tensor ring completion algorithms used for solving \eqref{eq:trals} such as alternating least-squares (ALS) can be applied to solve \eqref{eq:online11}. However, alternating minimization may incur a high computational cost, which is not computationally efficient for {streaming} processing. In this work, based on the fact that the tensor subspace within similar patches changes slowly (as Fig. \ref{fig:insight} shows), we put forth a one-step SSD method to obtain the update of the core tensors.

First, $\mathcal{Z}_k^t,k=1,\ldots,N-1$ are initialized with the corresponding previous estimates $\mathcal{Z}_k^{t-1}$. Then, the scaled gradient descent direction $g(\mathbf{Z}_{k(2)}^t)$ in terms of $\mathbf{Z}_{k(2)}^t$ is computed by first obtaining the gradient
\begin{equation}
\begin{aligned}
\nabla \mathcal{L}(\mathbf{Z}_{k(2)}^t)=-\left({\mathbf{P}^t_{[N]}}\!\circ\!({\mathbf{M}^t_{[N]}} -\mathbf{Z}_{k(2)}^t\mathbf{U}_k^{T})\right)\mathbf{U}_k
\end{aligned}
\label{eq:sgdupdate}
\end{equation}
where $\mathbf{U}_k=(\mathbf{Z}^{t})_{[2]}^{\neq k}$, and then adding a scaled term $\mathbf{U}_k^{T}\mathbf{U}_k$ such that
\begin{equation}
g(\mathbf{Z}_{k(2)}^t)=\nabla \mathcal{L}(\mathbf{Z}_{k(2)}^t)(\mathbf{U}_k^{T}\mathbf{U}_k+\epsilon\mathbf{I})^{-1}\:.
\label{eq:ssdg}
\end{equation}
where $\epsilon$ is a sufficient small positive number (i.e., $10^{-10}$). Therefore, $\mathbf{Z}_{k(2)}^t$ can be updated as
\begin{equation}
\mathbf{Z}_{k(2)}^t=\mathbf{Z}_{k(2)}^{t}-\mu_k g(\mathbf{Z}_{k(2)}^t)\:,
\label{eq:ssdupdate}
\end{equation}
where the step size $\mu_k$ is set by using exact line-search as
\begin{equation}
\begin{aligned}
\mu_k&\!=\!\arg\min_{\mu}{\left\|{{\mathbf{P}^t_{[N]}} \!\circ\! \left({\mathbf{M}^t_{[N]}}\!-\!(\mathbf{Z}_{k(2)}^t\!-\!\mu g(\mathbf{Z}_{k(2)}^t))\mathbf{U}_k^{T}\right)}\right\|_{F}^2}\\
&=-\frac{\langle\nabla \mathcal{L}(\mathbf{Z}_{k(2)}^t),g(\mathbf{Z}_{k(2)}^t)\rangle}{\left\|{\mathbf{P}^t_{[N]}} \circ \left(g(\mathbf{Z}_{k(2)}^t)\mathbf{U}_k^{T}\right)\right\|_F^2}\:.
\end{aligned}
\end{equation}
Compared with SGD, the SSD method increases the speed of convergence by using the scaled term $\mathbf{U}_k^{T}\mathbf{U}_k$ and applying an adaptive step size strategy. The above {streaming} tensor ring completion (STRC) algorithm is summarized in Algorithm 2.

\begin{algorithm}
\caption{{Streaming} tensor ring completion (STRC)}
\begin{algorithmic}[1]
 \REQUIRE Partially observed tensor $\mathcal{M}^t$ with corresponding mask tensor $\mathcal{P}^t$ and previous estimation $\mathcal{Z}_k^{t-1},k=1,\ldots,N-1$. Parameter $\gamma$.
 \STATE Compute $\mathcal{V}^t$ using \eqref{eq:online5} and set $\mathcal{Z}_N^t=\mathcal{V}^t$.
 \FOR {$k=1,\ldots,N-1$}
 \STATE Update $\mathcal{Z}_k^t$ using \eqref{eq:ssdupdate}.
 \ENDFOR
\ENSURE Updated core tensors $\mathcal{Z}_k^t,k=1,\ldots,N$.
\end{algorithmic}
\label{alg:STRC}
\end{algorithm}

\subsection{New patch tensor rank estimation and completion}

For a new patch tensor $\mathcal{M}^t\in\mathbb{R}^{I_1 \times \ldots \times I_{N-1}\times K_b}$ with corresponding mask tensor $\mathcal{P}^t$, there are no previous estimates of core tensors $\mathcal{Z}_k^{t-1},k=1,\ldots,N$. Therefore, one should apply a {batch} completion algorithm for $\mathcal{M}^t$ and create core tensors $\mathcal{Z}_k^{t}$. Inspired by the SSD method for STRC, we develop a new batch tensor ring completion algorithm using scaled steepest descent method (SSD). The method alternatively updates the core tensors $\mathcal{Z}_k^{t},k=1,\ldots,N$ using \eqref{eq:ssdupdate}.

Before completion and before obtaining the core tensors of a new patch tensor, we should estimate the TR rank of the tensor. {Following \cite{wang2017efficient,yuan2019tensor,huang2020provable,yu2020low}, we set all the elements of the TR rank equal to the same value, i.e. $r_1=\cdots=r_N=r$. Then, defining $p=\|\mathcal{P}^t\|_0/(\prod_{i=1}^{N-1}I_iK_b)$ as the observation ratio of the tensor $\mathcal{M}^t$, according to \cite[Theorem 1]{huang2020provable}, we consider setting { $r=C\sqrt{p}$, where $C$ is a free parameter. In practice, if the patch is texture-less, the rank should be set to a relatively small value for better completion performance.} Therefore, we propose the following adaptive strategy to estimate $r$:}
\begin{equation}
\begin{aligned}
r=\min(\lceil C_1\sqrt{p}\sigma^2_{\mathcal{M}_{\Omega}}\rceil,\lfloor C_2\sqrt{p}\rfloor+r_o)
\end{aligned}
\label{eq:rank}
\end{equation}
where $\sigma^2_{\mathcal{M}_{\Omega}}$ denotes the variance of the observed entries of $\mathcal{M}^t$, $p$ the observation ratio of $\mathcal{M}^t$. $C_1$, $C_2$ and $r_{o}$ are parameters to be set. 

We term the method proposed above TRSSD, and the pseudo code is summarized in Algorithm 3. The method proposed can outperform least-square-based method for patch tensor completion, which is verified in the experimental results.

\begin{algorithm}
\caption{Tensor ring completion with rank estimation and scaled steepest descent (TRSSD)}
\begin{algorithmic}[1]
 \REQUIRE Partially observed tensor $\mathcal{M}$ with corresponding mask $\mathcal{P}$, maximum iteration number $L$ and error tolerance $\varepsilon$.
 \STATE Set $r_1=\cdots=r_N=r$ and estimate $r$ using \eqref{eq:rank}.
 \STATE Initialize $\mathcal{Z}_{k},k=1,\ldots,N$. Set $\tau=1$.
 \REPEAT
 \FOR {$k=1,\ldots,N$}
 \STATE Update $\mathcal{Z}_k^{\tau}$ using \eqref{eq:ssdupdate}.
 \STATE $\tau=\tau+1$.
 \ENDFOR
 \STATE Compute $\mathbf{X}^{\tau}=\Re(\mathcal{Z}_1^{\tau},\mathcal{Z}_2^{\tau},\ldots,\mathcal{Z}_N^{\tau})$.
 \UNTIL {$\tau>L$ or $\|\mathbf{X}^{\tau}-\mathbf{X}^{\tau-1}\|_F/\|\mathbf{X}^{\tau-1}\|_F<\varepsilon$}
\ENSURE Core tensors $\mathcal{Z}_k,k=1,\ldots,N$.
\end{algorithmic}
\label{alg:TRBCSD}
\end{algorithm}

\section{Complexity analysis}
\label{sec:complexity}

{Without loss of generality, we assume that $m$ and $l$ are divisible by $s$, and the observation ratio for all patch tensors is the same value $p$. We first analyze the complexity of the proposed ECPM method. In particular, according to Section \ref{sec:onlinetrc}.B.2, the complexity of patch matching for each down-sampled patch is $\mathcal{O}(m^2nl^2/s^4)$. Therefore, the complexity of patch matching for a patch on all $s^2$ down-sampled frames is $\mathcal{O}(m^2nl^2/s^2)$. Suppose the total number of tracked patches for recovering the frame is $T$, then the overall complexity for ECPM on all tracked patches is $\mathcal{O}(m^2nl^2/s^2T)$. Compared with the traditional patch matching method (i.e., $s=1$), the proposed ECPM method can reduce the computational complexity by a factor of $s^2$. Moreover, since both patch matching on $T$ patches and coarse-scale matching for each patch on $s^2$ down-sampled frames are independent, the complexity can be further reduced by parallel computation.}

Next we analyze the computational complexity of patch tensor completion. The time complexity of STRC and for a tracked patch tensor of size $m\times m\times n\times K_o$ is $\mathcal{O}(pm^2nK_or^4)$, while TRSSD for a new patch tensor of size $m\times m\times n\times K_b$ requires time complexity of $\mathcal{O}(pm^2nK_br^4L)$. Define $T_1$ and $T_2$ as the number of tracked patch tensors and new patch tensors, respectively. Then, the overall complexity for tensor completion step is $\mathcal{O}(pm^2nK_or^4T_1+pm^2nK_br^4LT_2)$. Moreover, since tensor completion for each patch tensor is independent, parallel computation can also be applied to further improve the computational efficiency. 

\section{Experimental results}
\label{sec:results}

In this section, we conduct experiments to verify the performance of the proposed methods. We compare to both existing {batch and streaming} tensor completion algorithms. The {streaming} methods include the CP-based algorithm OLSTEC \cite{kasai2016online} and the t-SVD-based algorithm TOUCAN \cite{gilman2020online}. For batch methods, we compare to traditional algorithms including the matrix factorization-based algorithm TMAC \cite{xu2015parallel}, tensor ring-based algorithms TRNN \cite{yu2019tensor} and PTRC \cite{yu2020low}, and t-SVD-based algorithms TNN \cite{zhang2016exact} and TCTF \cite{zhou2017tensor}. We also compare two non-local (NL) patch-based batch methods including NLTNN \cite{zhang2019nonlocal} and tensor train-based NLTTNN \cite{ding2021tensor}.

The completion performance is evaluated by the average peak signal-to-noise ratio (PSNR) over 20 Monte Carlo runs with different missing entries. For all patch-based algorithms PTSTRC, NLTNN and NLTTNN, the patch size $m$ and the number of overlap pixels $o$ are set to $36$ and $12$, respectively. The number of nearest neighbors $K$ for NLTNN and NLTTNN is set to $20$, while $K_b$ and $K_o$ for PTSTRC are set to $30$ and $10$, respectively. The search size $l$ for PTSTRC and NLTNN are both set to $41$. The interval sampling parameter $s$, padding pixels $b$, threshold $\tau_f$ and $\tau_c$ for PTSTRC are set to $3$, $20$, $0.02$ and $3$ respectively. For the proposed STRC and TRSSD, the parameters $\gamma$, $L$, $C_1$, $C_2$, $r_o$ and $\varepsilon$ are set to $10^{-5}$, $10$, $1000$, $6$, $4$ and $10^{-2}$, respectively. For the other algorithms, the parameters are adjusted so as to achieve their best performance. Further, the parameters are fixed in each simulation. {All experiments were performed using MATLAB R2022a on a desktop PC with an 8-core 2.5-GHz processor i9-11900 and 32GB of RAM. Unless stated otherwise, parallel computation is applied to all patch-based methods for both patch tracking and tensor completion by using the MATLAB parallel computing toolbox on 8 CPU cores.}

\subsection{Ablation experiment}
{To verify the effectiveness of the proposed patch tracking and tensor completion methods, in this section we carry out ablation experiments with different algorithm settings. The experiment is carried out on a `water sport' video from the Moments in Time Dataset\footnote{\url{http://moments.csail.mit.edu}} \cite{monfortmoments}. Some representative frames are shown in the first row of Fig. \ref{fig:video_opticalflow}. The first $50$ frames of the video are selected, each frame with size $256\times342\times3$, resulting in a video tensor of size $256\times342\times3\times50$. The frames are assumed to arrive sequentially in a streaming fashion. The observation ratio is set to $p=0.2$, i.e., the observed pixels are selected uniformly at random with probability $0.2$. For fair comparison, parallel computation is not applied to patch tracking in this experiment.}

\begin{figure}[tb]
\centering
\includegraphics[width=1\linewidth]{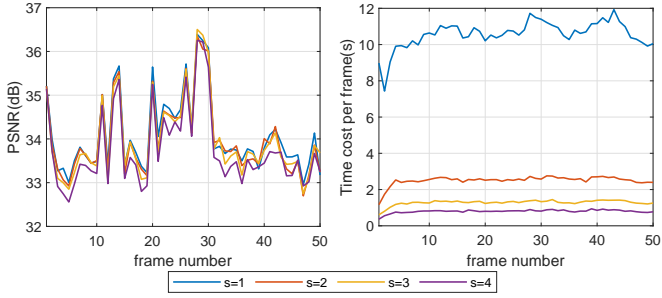}
\caption{Left: curves of average PSNR with different interval $s$. Right: average time cost of patch tracking per frame.}
\label{fig:ablation_interval}
\end{figure}

First, we verify the efficiency of the proposed ECPM method. Fig. \ref{fig:ablation_interval} depicts the PSNR of the recovered video for different values of interval parameter $s$, along with the computation time of patch matching for each frame. The dilation is utilized as default. Note that $s=1$ designates patch matching on the full-scale. As can be seen, using the proposed ECPM method ($s>1$) achieves significant speedup compared with full-scale matching. Further, as $s$ increases, the matching accuracy decreases, underscoring a tradeoff between matching accuracy and efficiency.

\begin{figure}[tb]
\centering
\vspace{-0.4cm}
\includegraphics[width=1\linewidth]{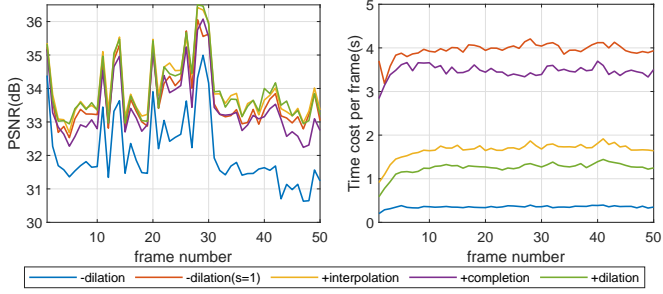}
\caption{Left: curves of average PSNR with different pre-processing methods ($s$ is set to $3$ as default); Right: average time cost of patch tracking per frame.}
\label{fig:ablation_dilation}
\vspace{-0.4cm}
\end{figure}

Second, we evaluate the performance with different pre-processing methods for patch matching with missing data, i.e., dilation, interpolation and completion. Fig. \ref{fig:ablation_dilation} depicts the PSNR and patch matching time cost under different pre-processing methods. The prefix `+' and `-' mean `with' and `without', respectively. {It should be noted that the result of $s=1$ without dilation corresponds to the traditional patch tracking method \cite{dabov2007image,ji2010robust,maggioni2012nonlocal,wen2017joint}}. One can observe that all methods achieve comparable good completion performance except for the one with $s=3$ without dilation, while the proposed dilation method with $s=3$ incurs a smaller computational cost for patch matching compared with interpolation and completion. Further, compared with Fig. \ref{fig:ablation_interval}, dilation greatly improves the patch matching accuracy, while increasing the computational cost. On the other hand, the speedup achieved due to coarse-scale patch matching counteracts the extra computational cost for matching using dilation. Therefore, our proposed patch tracking method can maintain both efficiency and accuracy. 

\begin{figure}[tb]
\centering
\includegraphics[width=1\linewidth]{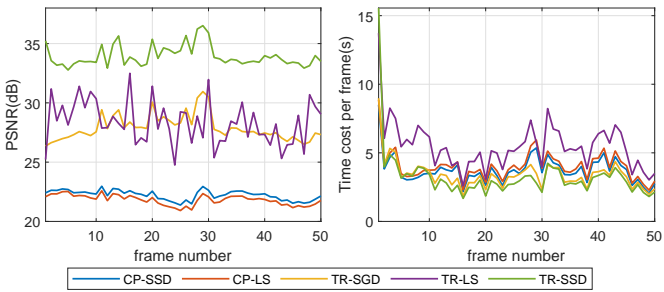}
\caption{Left: curves of average PSNR with different completion methods; Right: average time cost of tensor completion per frame.}
\label{fig:ablation_tcmethod}
\end{figure}

Third, we test performance using different tensor ring completion methods. For comparison, we developed two extra {streaming} tensor ring completion methods TR-SGD and TR-LS. The TR-SGD method utilizes the gradient in \eqref{eq:sgdupdate} with exact line-search for updating $\{\mathcal{Z}_{i}\}_{i=1}^{N-1}$. The least-squares method TR-LS applies one-step LS to update $\{\mathcal{Z}_{i}\}_{i=1}^{N-1}$. To compare with the CP-based algorithm, we also develop two methods CP-SSD and CP-LS by introducing SSD and LS to the CP model, respectively. The PSNR and completion time cost with different completion methods are shown in Fig. \ref{fig:ablation_tcmethod}. As shown, the proposed SSD method using the tensor ring model achieves the best completion performance, with computational cost smaller than that of TR-LS.

\begin{figure}[tb]
\centering
\includegraphics[width=1\linewidth]{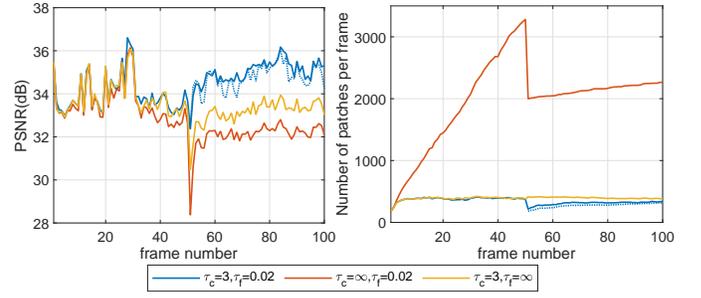}
\caption{Left: curves of average PSNR with different thresholds $\tau_c$ and $\tau_f$; Right: total number of patches for tensor completion per frame. Dotted blue line: results of completion using $\tau_c=3,\tau_f=0.02$ started from $51^\text{st}$ frame. }
\label{fig:ablation_overlap}
\end{figure}

Finally, we verify the effectiveness of pruning mistracked or highly overlapping patches. The test video is constructed by concatenating two videos (D$1$ and S$3$ in Fig. \ref{fig:video_opticalflow}), thus there exists a sudden scene change at the $51^\text{st}$ frame. Similar to the previous experiments, the observed pixels are selected uniformly at random with probability $0.2$. The PSNR and the total number of patches (sum of tracked patches and new patches) for completion are shown in Fig. \ref{fig:ablation_overlap} for different parameter settings. The $\tau_c=\infty$ denotes the approach without pruning overlapping patches, while $\tau_f=\infty$ is without pruning mistracked patches. One can observe that pruning mistracked and highly overlapping patches significantly improves the completion performance when the scene changes. Specifically, pruning overlapping patches significantly suppresses the growth of the number of patches, while pruning mistracked patches greatly improves the completion performance and reduces the number of patches when the scene changes. For comparison, we also show the completion results on the second video S$3$ alone (dotted blue line in Fig. \ref{fig:ablation_overlap}), i.e., the completion starts from the $51^\text{st}$ frame. As shown, with the proposed pruning strategies, the completion performance is not affected when the scene changes.

\subsection{Short video completion}
\begin{figure}[tb]
\centering
\includegraphics[width=1\linewidth]{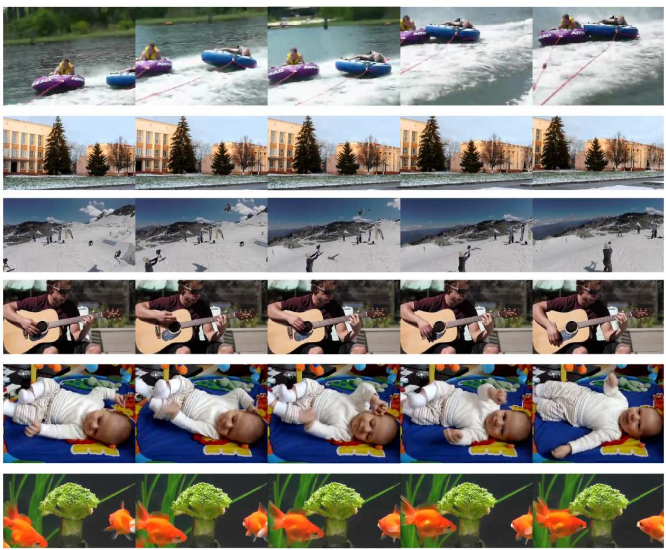}
\vspace{-0.6cm}
\caption{Representative frames of six videos. Frame number (from left to right): 5, 15, 25, 35 and 45. Video indices (from top to bottom): D1, D5, D18, S3, S18 and S30.}
\label{fig:video_opticalflow}
\end{figure}

In this part, we evaluate the completion performance using $50$ short color videos from the Moments in Time Dataset \cite{monfortmoments}. The Moments in Time Dataset is a large-scale short video dataset initially developed for recognizing and understanding actions in videos. In our experiment, $50$ videos from the dataset are randomly selected for performance test. Specifically, we randomly select $25$ videos with dynamic camera (i.e., the camera moves during the shoot) and $25$ videos with static camera (i.e., the camera does not move during the shoot). The dynamic and static camera videos are marked as D$1$-D$25$ and S$1$-S$25$, respectively. Representative frames from selected videos are shown in Fig. \ref{fig:video_opticalflow}. In general, dynamic camera videos always contain large motion across the video frames.

{Considering the computational limit of the batch tensor completion methods, for each video the first $50$ frames are selected such that the whole tensor is of size $I_1 \times I_2 \times 3 \times 50$. If the height is larger than $256$, the video is scaled down to a height of $256$. Frames arrive sequentially for TOUCAN, OLSTEC and PTSTRC, while for batch methods, the whole video tensor with all $50$ frames is given as the input tensor. {For PTRC and TRNN (which favor high order tensors for better performance), we reshape the tensor to a $9/10/11$-order tensor according to the factorization of $I_1$ and $I_2$ \cite{yu2020low}}. For TNN and TCTF, which only afford $3$-order tensor completion, we reshape the tensor to a $3$-order tensor of size $I_1\times I_2\times150$.}

\begin{table*}[htbp]
\caption{Average PSNR on six video groups with different missing patterns}
\label{tab:video}
\centering
\resizebox{1\textwidth}{!}{%
\begin{tabular}{cccccccccccccc}
\toprule
\multirow{2}{*}{\begin{tabular}[c]{@{}c@{}}Camera\\ Mode\end{tabular}} & \multirow{2}{*}{\begin{tabular}[c]{@{}c@{}}Video\\ index\end{tabular}} & \multirow{2}{*}{\begin{tabular}[c]{@{}c@{}}Missing\\ pattern\end{tabular}} & \multirow{2}{*}{\begin{tabular}[c]{@{}c@{}}parameter\\ $p$\end{tabular}} & \multicolumn{7}{c}{batch methods}                                       & \multicolumn{3}{c}{streaming methods} \\ \cmidrule(l){5-11} \cmidrule(l){12-14}
                                                                       &                                                                        &                                                                            &                                                                          & TMAC  & TNN   & TCTF  & TRNN  & PTRC  & NLTNN          & NLTTNN         & TOUCAN  & OLSTEC  & PTSTRC           \\ \midrule
\multirow{9}{*}{Dynamic}                                               & \multirow{3}{*}{D1-D5}                                                 & \multirow{3}{*}{\begin{tabular}[c]{@{}c@{}}random\\ pixel\end{tabular}}   & 0.1                                                                      & 20.72 & 23.89 & 19.67 & 24.71 & 25.51 & 26.93          & {\ul 28.39}    & 20.46    & 21.66    & \textbf{31.63}  \\
                                                                       &                                                                        &                                                                            & 0.2                                                                      & 28.83 & 27.18 & 21.29 & 28.58 & 28.48 & 29.62          & {\ul 32.41}    & 22.70    & 24.71    & \textbf{35.35}  \\
                                                                       &                                                                        &                                                                            & 0.3                                                                      & 32.02 & 29.39 & 22.11 & 31.83 & 30.16 & 31.66          & {\ul 35.57}    & 24.39    & 26.46    & \textbf{37.82}  \\ \cmidrule(l){2-14} 
                                                                       & \multirow{3}{*}{D6-D10}                                                & \multirow{3}{*}{\begin{tabular}[c]{@{}c@{}}random\\ stripe\end{tabular}}  & 0.1                                                                      & 12.33 & 24.03 & 16.60 & 23.09 & 18.08 & 22.64          & {\ul 25.77}    & 20.41    & 19.28    & \textbf{28.31}  \\
                                                                       &                                                                        &                                                                            & 0.2                                                                      & 22.42 & 27.57 & 21.11 & 27.84 & 26.62 & 26.37          & {\ul 31.69}    & 22.78    & 23.33    & \textbf{34.15}  \\
                                                                       &                                                                        &                                                                            & 0.3                                                                      & 26.40 & 30.69 & 22.66 & 31.95 & 30.02 & 29.15          & {\ul 35.56}    & 24.60    & 26.21    & \textbf{37.77}  \\ \cmidrule(l){2-14} 
                                                                       & \multirow{3}{*}{D11-D15}                                               & \multirow{3}{*}{\begin{tabular}[c]{@{}c@{}}random\\ tube\end{tabular}}    & 0.1                                                                      & 7.67  & 18.82 & 10.90 & 18.54 & 17.71 & \textbf{24.05} & 16.49          & 14.20    & 18.30    & {\ul 23.99}     \\
                                                                       &                                                                        &                                                                            & 0.2                                                                      & 10.04 & 22.37 & 15.76 & 22.58 & 20.26 & {\ul 26.84}    & 25.44          & 16.38    & 21.55    & \textbf{27.87}  \\
                                                                       &                                                                        &                                                                            & 0.3                                                                      & 13.77 & 25.09 & 18.45 & 25.48 & 22.36 & 28.75          & {\ul 29.29}    & 17.92    & 24.24    & \textbf{30.41}  \\ \midrule
\multirow{9}{*}{Static}                                                & \multirow{3}{*}{S1-S5}                                                 & \multirow{3}{*}{\begin{tabular}[c]{@{}c@{}}random\\ pixel\end{tabular}}   & 0.1                                                                      & 22.85 & 26.20 & 21.07 & 25.57 & 23.26 & 25.49          & \textbf{31.80} & 21.91    & 21.05    & {\ul 30.40}     \\
                                                                       &                                                                        &                                                                            & 0.2                                                                      & 30.41 & 29.36 & 22.31 & 29.57 & 28.25 & 28.05          & \textbf{35.79} & 24.02    & 24.41    & {\ul 34.24}     \\
                                                                       &                                                                        &                                                                            & 0.3                                                                      & 31.85 & 31.67 & 23.17 & 32.62 & 30.17 & 29.95          & \textbf{38.79} & 25.54    & 26.78    & {\ul 36.82}     \\ \cmidrule(l){2-14} 
                                                                       & \multirow{3}{*}{S6-S10}                                                & \multirow{3}{*}{\begin{tabular}[c]{@{}c@{}}random\\ stripe\end{tabular}}  & 0.1                                                                      & 11.41 & 27.13 & 16.07 & 24.34 & 17.53 & 20.89          & \textbf{28.85} & 20.70    & 17.71    & {\ul 27.44}     \\
                                                                       &                                                                        &                                                                            & 0.2                                                                      & 20.14 & 31.12 & 21.08 & 30.16 & 28.32 & 24.25          & \textbf{37.27} & 22.58    & 22.64    & {\ul 34.01}     \\
                                                                       &                                                                        &                                                                            & 0.3                                                                      & 22.68 & 34.31 & 21.90 & 34.69 & 31.09 & 27.16          & \textbf{41.03} & 24.02    & 24.99    & {\ul 37.90}     \\ \cmidrule(l){2-14} 
                                                                       & \multirow{3}{*}{S11-S15}                                               & \multirow{3}{*}{\begin{tabular}[c]{@{}c@{}}random\\ tube\end{tabular}}    & 0.1                                                                      & 7.55  & 18.30 & 11.13 & 17.70 & 18.84 & {\ul 21.70}    & 14.51          & 9.52     & 18.02    & \textbf{21.85}  \\
                                                                       &                                                                        &                                                                            & 0.2                                                                      & 9.53  & 21.69 & 17.21 & 21.02 & 21.51 & {\ul 23.93}    & 22.96          & 12.08    & 20.92    & \textbf{25.24}  \\
                                                                       &                                                                        &                                                                            & 0.3                                                                      & 12.11 & 24.05 & 20.28 & 23.54 & 23.38 & 25.55          & {\ul 26.92}    & 8.26    & 22.91    & \textbf{27.96}  \\ \bottomrule
\end{tabular}%
}
\end{table*}

\begin{figure}[htbp]
\centering
\includegraphics[width=1\linewidth]{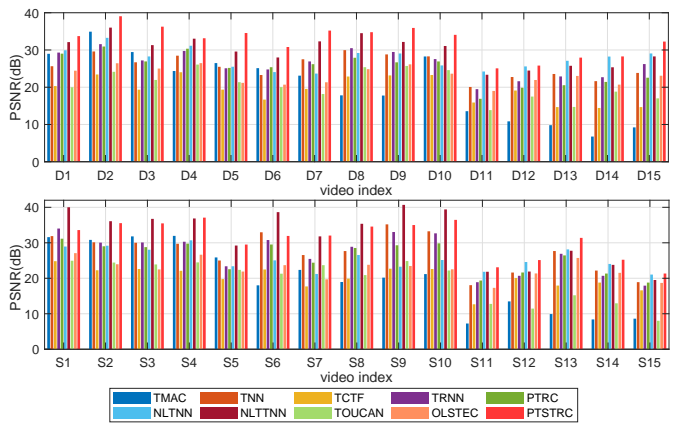}\\
\caption{Average PSNR for each video using different algorithms corresponding to Table \ref{tab:video} under $p=0.2$. Top: dynamic camera videos D1-D15 ; Bottom: static camera videos S1-S15. }
\label{fig:video_psnr}
\end{figure}

\begin{figure}[htbp]
\centering
\includegraphics[width=1\linewidth]{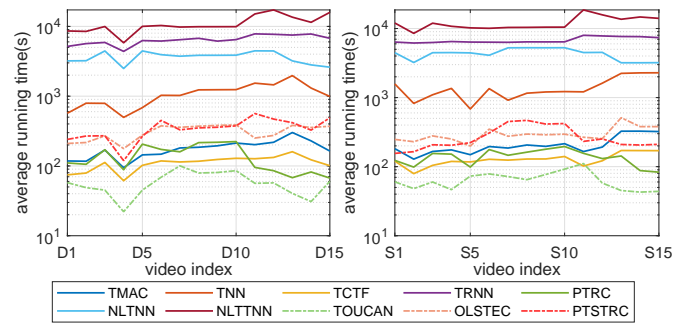}
\caption{Average running time for each video corresponding to Fig.\ref{fig:video_psnr} ($p=0.2$). Left: dynamic camera videos D1-D15 ; Right: static camera videos S1-S15.}
\label{fig:video_time}
\end{figure}

\begin{figure*}[htbp]
\centering
\includegraphics[width=1\linewidth]{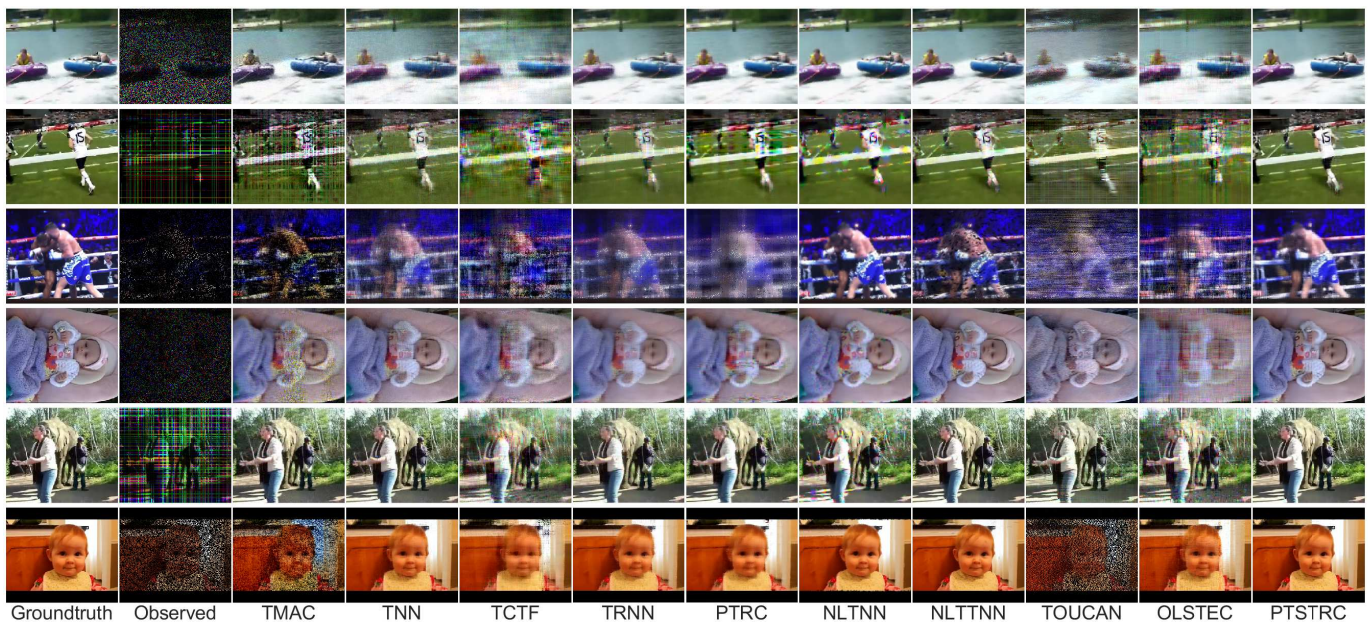}
\caption{Examples of recovered $25^\text{th}$ frames of six videos corresponding to Table \ref{tab:video}. Video indices with $p$ settings (from top to bottom): D1 ($p=0.2$), D6 ($p=0.1$), D11 ($p=0.1$), S2 ($p=0.1$), S7 ($p=0.2$), S12 ($p=0.3$). For better view, frames may be cropped on both sides.}
\label{fig:video_example}
\end{figure*}

\begin{table}[htbp]
\caption{Average PSNR on four video groups with different missing patterns}
\label{tab:video2}
\centering
\resizebox{0.47\textwidth}{!}{%
\begin{tabular}{@{}cccccc@{}}
\toprule
\multicolumn{2}{c}{Camera mode}                                                       & \multicolumn{2}{c}{Dynamic}                                                                                           & \multicolumn{2}{c}{Static}                                                                                            \\ \cmidrule(r){1-2} \cmidrule(r){3-4} \cmidrule(r){5-6}
\multicolumn{2}{c}{\begin{tabular}[c]{@{}c@{}}Video\\ index\end{tabular}}             & D16-D20                                                 & D21-D25                                                     & S16-S20                                                 & S21-S25                                                     \\ \midrule
\multicolumn{2}{c}{\begin{tabular}[c]{@{}c@{}}Missing\\ pattern\end{tabular}}         & \begin{tabular}[c]{@{}c@{}}random\\ block\end{tabular} & \begin{tabular}[c]{@{}c@{}}fixed\\ watermark\end{tabular} & \begin{tabular}[c]{@{}c@{}}random\\ block\end{tabular} & \begin{tabular}[c]{@{}c@{}}fixed\\ watermark\end{tabular} \\ \midrule
\multirow{7}{*}{\begin{tabular}[c]{@{}c@{}}batch \\ methods\end{tabular}}    & TMAC   & 19.40                                                   & 19.68                                                       & 18.91                                                   & 18.38                                                       \\
                                                                             & TNN    & 28.25                                                   & 32.83                                                       & {\ul 37.54}                                             & 31.73                                                       \\
                                                                             & TCTF   & 20.80                                                   & 20.10                                                       & 20.15                                                   & 22.82                                                       \\
                                                                             & TRNN   & 28.07                                                   & 34.28                                                       & 34.83                                                   & 32.38                                                       \\
                                                                             & PTRC   & 27.10                                                   & 30.19                                                       & 33.12                                                   & 25.59                                                       \\
                                                                             & NLTNN  & 27.65                                                   & 36.11                                                       & 27.90                                                   & 34.25                                                       \\
                                                                             & NLTTNN & 30.31                                                   & {\ul 36.98}                                                 & \textbf{38.76}                                          & {\ul 34.76}                                                 \\ \midrule
\multirow{2}{*}{\begin{tabular}[c]{@{}c@{}}learning\\ methods\end{tabular}}  & STTN   & 29.76                                                   & 30.75                                                       & 32.61                                                   & 29.92                                                       \\
                                                                             & ViF    & {\ul 30.47}                                             & 32.21                                                       & 32.19                                                   & 29.99                                                       \\ \midrule
\multirow{3}{*}{\begin{tabular}[c]{@{}c@{}}streaming\\ methods\end{tabular}} & TOUCAN & 22.62                                                   & 24.43                                                       & 26.09                                                   & 20.57                                                       \\
                                                                             & OLSTEC & 21.19                                                   & 27.45                                                       & 23.97                                                   & 30.13                                                       \\
                                                                             & PTSTRC & \textbf{30.71}                                          & \textbf{38.59}                                              & 34.79                                                   & \textbf{35.28}                                              \\ \bottomrule
\end{tabular}%
}
\end{table}

\begin{figure}[htbp]
\centering
\includegraphics[width=1\linewidth]{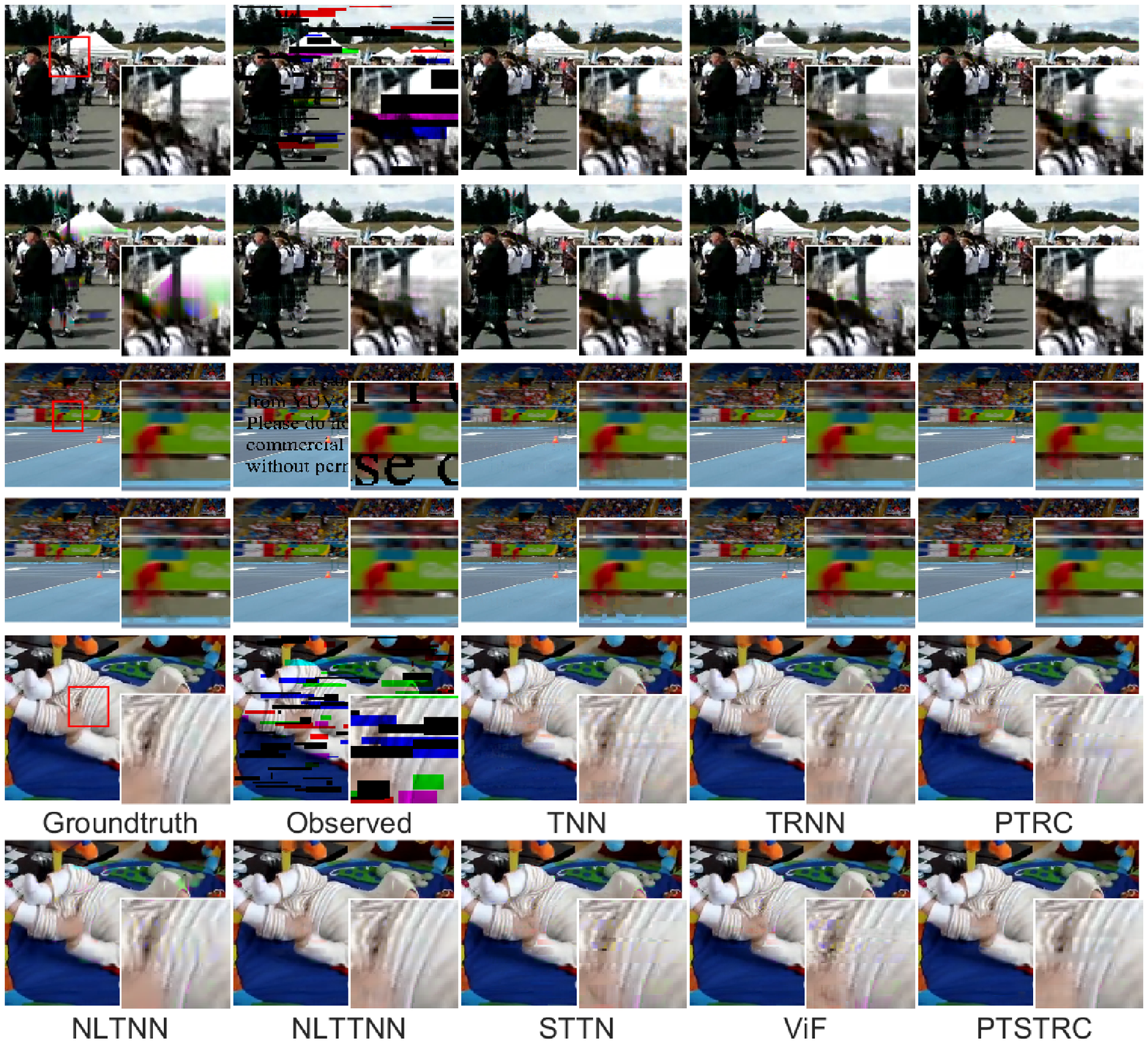}
\vspace{-0.6cm}
\caption{Examples of recovered $25^\text{th}$ frames of three videos corresponding to Table \ref{tab:video2} for partial algorithms. Video indices (from top to bottom): D19, D25 and S18. Bottom-right window of each frame shows the enlarged view of specific region (red rectangles in the groundtruth frames).}
\label{fig:video_example2}
\end{figure}

{First, we investigate the performance of all algorithms on three missing data patterns commonly used in tensor completion performance verification, namely, random pixel/stripe/tube. 
In particular, in random pixel, the pixels are observed uniformly at random with observation ratio $p\times100\%$. For the random stripe missing pattern, the observed columns and rows are randomly and uniformly selected with probability $p\times100\%$. For random tube, a time-invariant random missing pattern with observation ratio $p\times100\%$ is applied to each frame, i.e., the locations of the observed pixels are the same for all $50$ frames.}

{Table \ref{tab:video} shows the average PSNR of different algorithms for dynamic and static camera videos. Each value of PSNR is averaged over a group of five videos. More detailed PSNR results for each video with $p=0.2$ are shown in Fig. \ref{fig:video_psnr}, and the corresponding average running times are shown in Fig. \ref{fig:video_time}. For better visualization, examples of recovered frames using different algorithms are shown in Fig. \ref{fig:video_example}. It can be observed that the proposed PTSTRC achieves the best performance for dynamic cases and second best for static cases. The superior performance of PTSTRC on dynamic videos confirms that the completion performance can be improved by leveraging temporal consistency within similar patches. Further, although NLTTNN outperforms PTSTRC on S$1$-S$10$, its computational cost is considerably larger than PTSTRC as shown in Fig. \ref{fig:video_time}. More importantly, NLTTNN is a batch method that needs to access all $50$ frames at one time, while our proposed PTSTRC is a streaming method that can sequentially complete each frame.}

{Second, we test the performance on two missing data patterns that are more common in practical applications. The first is the random block missing pattern, which could arise in video recovery with packet loss \cite{feamster2002packet}. In particular, for each frame, a number $N\in[50,150]$ of blocks with height $H\in[1,10]$ and width $W\in[1,100]$ are randomly generated and selected. In each block, partial color channel or all color channels are missing. The second is the watermark missing pattern which is directly related to the application of watermark removal. Specifically, in our experiment a fixed sentence is masked on all frames. Examples of some observed frames are shown in Fig. \ref{fig:video_example2}.

We also include two popular learning-based video inpainting methods, Spatial-Temporal Transformer Network (STTN) \cite{zeng2020learning} and Video inpainting with Fuseformer (ViF) \cite{liu2021fuseformer}, for comparison. Table \ref{tab:video2} shows the average PSNR for each video group for all algorithms, and examples of recovered frames for partial algorithms are shown in Fig. \ref{fig:video_example2}. TMAC, TCTF, TOUCAN and OLSTEC are not shown due to limited space. It can be observed that the proposed PTSTRC achieves the best overall performance with better visualization on dynamic videos. TNN or NLTTNN work better than PTSTRC on the random block missing pattern under static videos (S$16$-S$20$). An example of a recovered frame of S$18$ is shown in Fig. \ref{fig:video_example2}. It can be observed that the smearing of PTSTRC on missing regions is more obvious than TNN or NLTTNN.}

\subsection{Long streaming video completion}
In this section, we verify the completion performance on a live basketball video downloaded from YouTube\footnote{\url{https://www.youtube.com/watch?v=I33o9UnUe1A&t=986s}}. We select $500$ consecutive frames and resize each frame to size $360\times 640$. The frames arrive sequentially in a {streaming} manner. For each frame, the observed pixels are randomly and uniformly selected with probability $20\%$. Since batch completion methods cannot handle such a large tensor, we only compare the performance to the {streaming} completion methods OLSTEC and TOUCAN. 

\begin{figure}[htbp]
\centering
\includegraphics[width=1\linewidth]{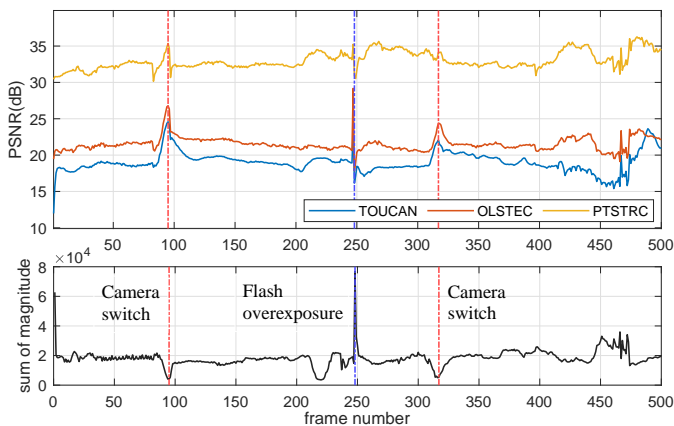}
\vspace{-0.6cm}
\caption{Top: curves of average PSNR for each frame using different algorithms on basketball streaming video. Bottom: curves of average magnitude of optical flow between frames of the original fully observed video. }
\label{fig:nba_500}
\end{figure}

\begin{figure}[htbp]
\centering
\includegraphics[width=1\linewidth]{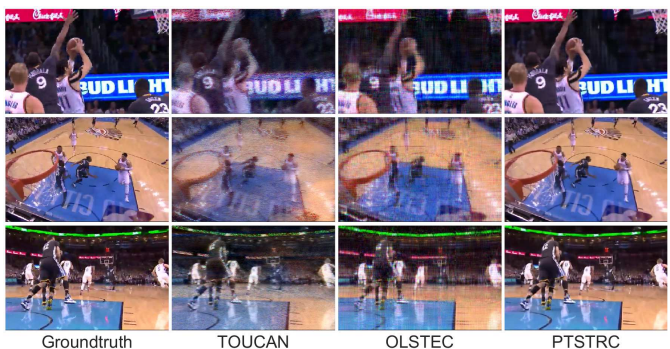}
\vspace{-0.6cm}
\caption{Example of recovered frames (middle part). From top to bottom: frame $50$, $200$ and $450$. Best viewed in $\times2$ sized color pdf file.}
\label{fig:nba_500_example}
\end{figure}

Fig. \ref{fig:nba_500} depicts the PSNR of each recovered frame, along with the sum of magnitude of the optical flow between frames. Some sudden changes have been marked in the figure including two camera switch events in which the whole scene changes, and one flash overexposure where the frame is almost entirely white. Some recovered frames using different algorithms are also displayed in Fig. \ref{fig:nba_500_example}. As shown, the proposed PTSTRC significantly outperforms OLSTEC and TOUCAN. 

\vspace{-0.3cm}

\subsection{Image (single frame) completion}
{In this part, we verify the performance of the proposed method on the task of image completion. As described in Remark 1, image completion can be regarded as a special case of streaming video completion where there is only one incoming frame at $t=1$. In this case, only the TRSSD algorithm is utilized, and patch tensors are constructed using patch matching (PM) within the image. To avoid ambiguity, we name PTSTRC under this setting patch matching-based TRSSD (PMTRSSD). According to Remark 1, we also develop two new algorithms, namely, PMTTNN and PMTRNN, where TRNN\cite{yu2019tensor} and TTNN\cite{bengua2017efficient} are utilized for patch tensor completion, respectively. The algorithm settings for PMTRSSD, PMTRNN and PMTTNN are the same as PTSTRC in the previous experiment.}

\begin{figure*}[htbp]
\centering
\includegraphics[width=1\linewidth]{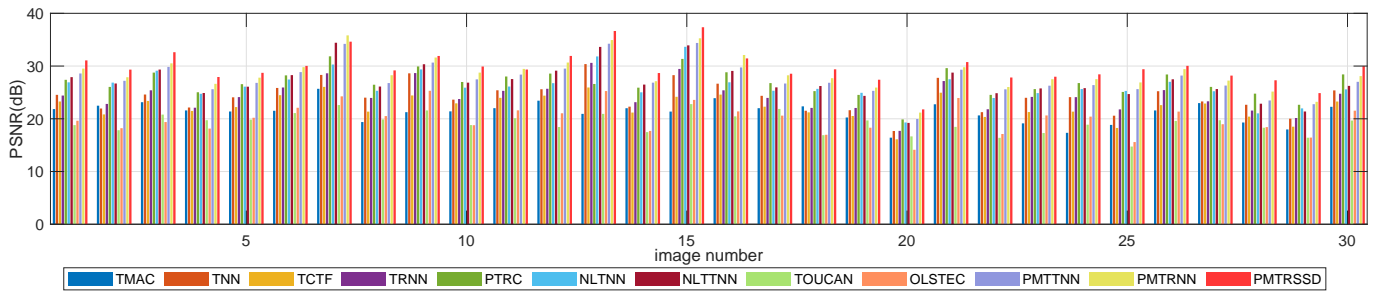}
\vspace{-0.6cm}
\caption{Comparison of the average PSNR for different algorithms on 30 selected test images with $p=0.2$.}
\label{fig:image_psnr}
\end{figure*}

\begin{figure}[htbp]
\centering
\includegraphics[width=1\linewidth]{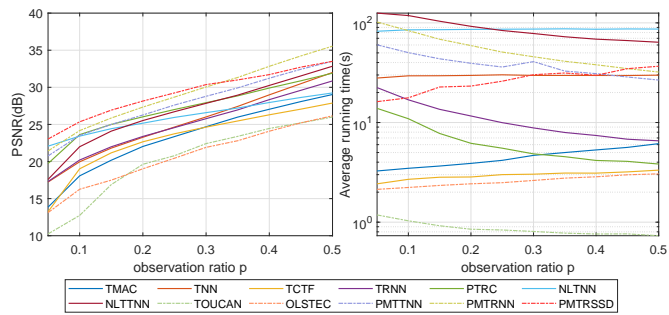}
\vspace{-0.6cm}
\caption{Left: average PSNR with different observation ratio $p$. Right: average time cost with different $p$.}
\label{fig:image_p}
\end{figure}

\begin{figure}[htbp]
\centering
\includegraphics[width=1\linewidth]{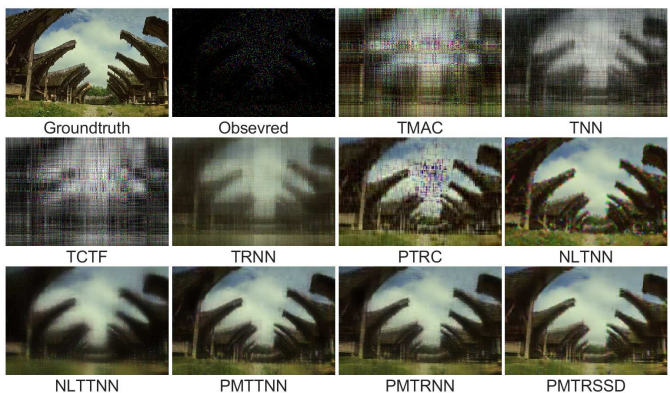}
\vspace{-0.7cm}
\caption{Example of recovered image in Fig. \ref{fig:image_p} with $p=0.05$ using different algorithms. Best viewed in $\times2$ sized color pdf file.}
\label{fig:image_example}
\end{figure}

We use the Berkeley Segment Dataset \cite{MartinFTM01} as our test data; $30$ images are randomly selected from the dataset and reshaped to a size of $320\times480$. For each image, $p\times100\%$ of the pixels are randomly and uniformly selected as the observed data. The image completion problem is then formulated as a $320\times480\times3$ tensor completion task. For PTRC and TRNN, we reshape the tensor to an $9$-order tensor of size $4\times4\times4\times5\times4\times4\times5\times6\times3$. In this experiment, the number of nearest neighbors $K$ for all patch-based method is set to $30$.

First, we compare the completion performance of all algorithms on all the $30$ test images. Fig. \ref{fig:image_psnr} shows the average PSNR for the test images with $p=0.2$. As shown, our proposed method achieves the best performance and PMTRNN comes second. The results verify that the patch-based method can improve performance upon methods that treat the entire image as a tensor. 

Second, we investigate the performance of the algorithms with different observation ratio $p$. Fig. \ref{fig:image_p} shows the completion performance for each algorithm for different values of $p$ for the image `bamboo house' (See Fig. \ref{fig:image_example}). As can be seen, patch-based tensor ring methods (PMTRSSD and PMTRNN) achieve the overall best performance for all $p$ compared with other methods. Moreover, for the low observability regime with $p<0.3$, PMTRSSD outperforms PMTRNN in PSNR with a significantly smaller computational cost. When the number of observed pixels is relatively large (i.e., $p>0.3$), PMTRNN has better performance than PMTRSSD. An example of recovered images under $p=0.05$ is also provided in Fig. \ref{fig:image_example}. As shown, the completed images using the proposed PMTRSSD and PMTRNN have better visual performance than other algorithms. It also shows that NLTNN suffers from color inconsistencies in some pixels due to misguidance from wrong interpolation. {We should remark that although PMTRNN outperforms PMTRSSD in some cases, as mentioned in Remark 1, PMTRNN cannot be adapted to a streaming setting and its application is limited to image completion.}

\begin{figure}[htbp]
\centering
\includegraphics[width=1\linewidth]{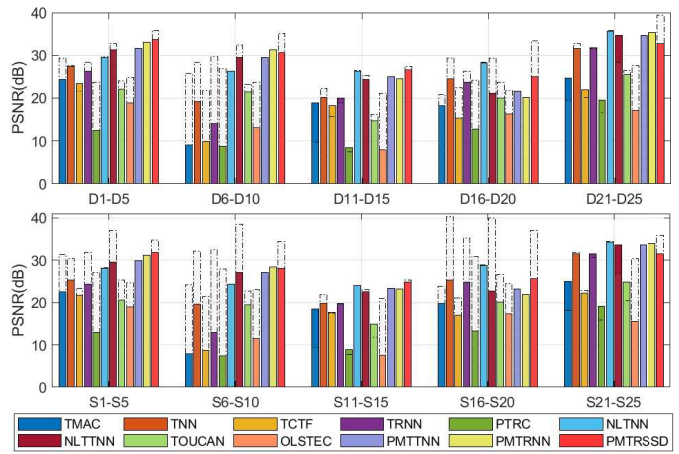}
\vspace{-0.6cm}
\caption{Solid bar: average PSNR for the $25^\text{th}$ frame of videos using different image completion algorithms. Dotted bar: the corresponding PSNR of the $25^\text{th}$ frame from previous video completion results. The dotted bar of PMTRSSD denotes the PSNR of PTSTRC. }
\label{fig:image_YUV}
\end{figure}

{For streaming tensor completion where the frames arrive sequentially, a batch method can be applied on a frame-by-frame basis, i.e., each incoming frame $\mathcal{Y}^t$ can be treated as a single image and completed individually. We carry out an experiment to investigate the performance when completion is done frame-by-frame for the video data. In particular, each $25^\text{th}$ frame of the $50$ short videos used in the previous subsection is completed by image completion algorithms. The missing patterns of these frames are the same as in the previous experiments ($p=0.2$ for D$1$-D$15$ and S$1$-S$15$). Fig. \ref{fig:image_YUV} depicts the average PSNR for completion of the $10$ groups corresponding to previous video groups. The dotted bars denote the corresponding average PSNR of the $25^\text{th}$ frames obtained from the previous video completion results. As can be seen, most of the algorithms suffer from some performance degradation with image completion (solid bars) compared to the corresponding results using video completion (dotted bars) where the adjacent frames were considered together. This experiment verifies that it is beneficial to exploit the relation between frames rather than treat each frame individually.}

\section{Conclusion}
\label{sec:conclusion}
In this paper, we proposed a new patch tracking-based {streaming} tensor ring completion (PTSTRC) framework for visual data recovery. The framework exploits the high correlation between patches to improve the completion performance. To improve the patch matching efficiency, an efficient coarse-scale patch matching (ECPM) method is proposed. A new {streaming} tensor ring completion (STRC) algorithm using scaled steepest descent is developed, which can simultaneously achieve favorable completion performance and high computational efficiency. Equipped with an efficient ECPM method and a fast STRC algorithm, the proposed algorithm can obtain better performance than the state-of-the-art batch and {streaming} tensor completion algorithms.


\ifCLASSOPTIONcaptionsoff
  \newpage
\fi

\bibliographystyle{IEEEtran}
\bibliography{references}

\end{document}